\documentclass[a4paper,fleqn]{cas-dc}

\usepackage{subcaption}
\usepackage{caption}
\usepackage{multicol}
\usepackage[numbers,sort&compress]{natbib}
\usepackage{amsmath,amsfonts,amssymb}
\usepackage{algorithmic}
\usepackage{algorithm}
\usepackage{array}
\usepackage{textcomp}
\usepackage{stfloats}
\usepackage{url}
\usepackage{verbatim}
\usepackage{graphicx}
\usepackage{float}
\usepackage{xcolor}
\usepackage{booktabs}
\usepackage{tabularx}
\usepackage{adjustbox}
\usepackage{hyperref}

\hypersetup{colorlinks=true, citecolor=blue, linkcolor=blue, urlcolor=blue}
\definecolor{text-red}{RGB}{255,0,0}
\definecolor{text-blue}{RGB}{0,0,255}
\newcommand{\rv}[1]{\textcolor{black}{#1}}

\shorttitle{Gao et al.}
\usepackage{placeins}
\usepackage{balance}
\usepackage{fancyhdr}
\usepackage{lastpage}
\makeatletter
\def\ps@headings{%
  \def\@oddfoot{\hfill\thepage\hfill}%
  \def\@evenfoot{\hfill\thepage\hfill}%
  \def\@oddhead{}%
  \def\@evenhead{}%
}
\makeatother

\begin{document}

\let\printorcid\relax

\title[mode=title]{DFIR-DETR: Frequency-Domain Iterative Refinement and Dynamic Feature Aggregation for Small Object Detection}

\author[1]{Bo~Gao}
\author[1]{Jingcheng~Tong}
\author[2]{Xingsheng~Chen}
\author[3]{Han~Yu}
\author[1]{Zichen~Li}

\affiliation[1]{organization={School of Information Engineering, Beijing Institute of Graphic Communication}, city={Beijing}, country={China}}

\affiliation[2]{organization={School of Computing and Data Science, The University of Hong Kong}, city={Hong Kong SAR}, country={China}}

\affiliation[3]{organization={College of Computing and Data Science, Nanyang Technological University}, country={Singapore}}

\begin{abstract}
Small object detection in complex scenes exposes a fundamental tension in neural network design: backbone attention distributes computation uniformly regardless of content, pyramid necks inflate activation magnitudes during upsampling without norm compensation, and bottleneck convolutions progressively smooth high-frequency edge components through accumulated spatial filtering. \rv{In response, we develop DFIR-DETR by tracing each proposed module back to a specific, measurable deficiency in the RT-DETR baseline: uniform attention that ignores spatial complexity, norm drift that destabilises upsampled features, and spatial convolutions that progressively suppress the high-frequency components small objects depend on.} On NEU-DET and VisDrone, DFIR-DETR achieves 92.9\% and 51.6\% mAP50 with only 11.7M parameters and 47.2~GFLOPs, demonstrating consistent gains across two qualitatively different detection domains.
\end{abstract}


\maketitle
\thispagestyle{fancy}
\pagestyle{fancy}
\fancyhf{}
\renewcommand{\headrulewidth}{0pt}
\renewcommand{\footrulewidth}{0pt}
\fancyfoot[R]{\footnotesize\thepage}
\section{Introduction}

Small object detection remains one of the most demanding problems in deep neural network research. Targets occupying fewer than $32\times32$ pixels carry limited texture information, forcing the network to rely heavily on boundary signals and long-range contextual cues that standard architectures are poorly equipped to exploit. \rv{The practical stakes differ sharply between the two application domains this work targets. On industrial production lines, a micro-defect missed during hot-rolled steel inspection does not stay missed; mechanical stress concentrates at the undetected boundary and the failure propagates downstream. Throughput constraints are equally unforgiving: frame budgets measured in tens of milliseconds rule out the multi-pass strategies that offline detectors rely on for accuracy. Aerial monitoring from UAVs presents a structurally different version of the same difficulty. At cruising altitude, a pedestrian or cyclist collapses to 10--15 pixels; the surrounding scene is dense and cluttered, so both missed targets and spurious detections carry real operational cost. What unites the two domains is a shared bottleneck: neither can tolerate the loss of high-frequency boundary information that deep spatial filtering progressively introduces. What separates them is everything else: background statistics, object formation mechanism, scale distribution, imaging geometry. A detector that holds up across both is harder to dismiss as a dataset-specific artefact.}

The difficulty is further compounded in cross-scene settings, where a single model must generalise across domains as different as UAV aerial imagery and industrial surface inspection.

A closer examination of modern real-time detectors reveals three structural limitations that bear directly on small object performance. First, convolutional backbones allocate attention computation uniformly across the spatial domain, giving equal weight to uninformative backgrounds and information-dense object boundaries~\cite{yuan2024small}. Second, feature pyramid necks expand activation magnitudes during upsampling without compensating normalisation, disrupting gradient dynamics and degrading cross-scale feature fusion~\cite{wang2024attention}. Third, repeated spatial convolutions act as implicit low-pass filters, attenuating the high-frequency edge components that small objects depend upon for accurate localisation~\cite{chi2020fast}.

RT-DETR~\cite{zhao2024rtdetr} represents the current state of the art in real-time transformer detection, combining a CNN backbone with a hybrid encoder and an end-to-end query based decoder. However, its ResNet backbone applies fixed-budget attention without content adaptation, its CCFF neck upsamples without amplitude control, and its RepC3 bottleneck operates entirely in the spatial domain, leaving all three limitations unaddressed.

We propose DFIR-DETR, which introduces three architectural contributions corresponding to each identified limitation. The Dynamic Content-Feature Aggregation (DCFA) module  parsifies the attention matrix through a dynamic Top-K mechanism predicted from local feature statistics, reducing complexity from $\mathcal{O}(N^2)$ to $\mathcal{O}(NK)$. The Dynamic Feature Pyramid Network (DFPN) introduces norm-preserving upsampling together with a dual-path convolution structure for explicit spatial detail recovery. The Frequency domain Iterative Refinement module (FIRC3) reformulates feature aggregation as a constrained optimisation problem in the spectral domain, where high-frequency boundary components can be explicitly reinforced.

The main contributions of this work are summarised as follows:

\begin{itemize}
    \item DCFA is introduced as a content-adaptive backbone module, where a dynamic Top-K sparsification mechanism redistributes attention capacity toward structurally complex regions, bringing complexity down from  $\mathcal{O}(N^2)$ to $\mathcal{O}(NK)$ without sacrificing global context modelling.

    \item Drawing on an $L_1$-norm conservation argument, DFPN is developed with analytically derived amplitude normalisation during upsampling and a dual-path shuffle convolution for explicit fine-grained spatial detail recovery across pyramid scales.

    \item The bottleneck feature aggregation problem is reformulated as a
    frequency-domain least-squares optimisation in FIRC3, granting the
    network direct and learnable access to high-frequency boundary components
    that accumulate spatial filtering cannot preserve.

    \item On NEU-DET and VisDrone, the proposed architecture attains 92.9\%
    and 51.6\% mAP50 with only 11.7M parameters and 47.2~GFLOPs, achieving
    state-of-the-art accuracy while reducing both model size and computational
    cost relative to the baseline.
\end{itemize}

The remainder of this paper is organised as follows.
Section~\ref{sec:related} reviews related work on object detection
architectures, attention mechanisms, multi-scale feature fusion, and
frequency-domain neural network modules. Section~\ref{sec:methods} presents
the proposed DFIR-DETR framework and details the design of DCFA, DFPN, and
FIRC3. Section~\ref{sec:experiments} describes the experimental setup,
benchmark datasets, ablation studies, and comparisons with state-of-the-art
methods. Section~\ref{sec:conclusion} concludes the paper.

\section{Related Work}
\label{sec:related}

\subsection{Object Detection Architectures}

Object detection has undergone two broad \linebreak paradigm shifts over the past decade.
Two-stage methods such as Faster R-CNN~\cite{ren2017faster} separated region
proposal from classification, achieving strong accuracy at the cost of
inference speed. One-stage detectors including SSD~\cite{liu2016ssd} and
YOLO~\cite{redmon2016yolo} collapsed this pipeline into a single forward pass,
with YOLOv3~\cite{redmon2018yolov3} demonstrating that multi-scale prediction
heads could recover much of the accuracy lost by removing the proposal stage.

DETR~\cite{carion2020detr} subsequently recast detection as set prediction
via bipartite matching, discarding both anchors and non-maximum suppression.
Deformable DETR~\cite{zhu2021deformable}, DN-DETR~\cite{li2022dn}, and
DINO~\cite{zhang2023dino} progressively addressed its convergence and accuracy
limitations. RT-DETR brought this line of work to
practical deployment by pairing a hybrid encoder with a query-based decoder,
achieving real-time throughput while retaining end-to-end training. The
present work takes RT-DETR as its starting point, leaving the decoder
unchanged while substantially rethinking how features are extracted, fused,
and refined prior to decoding.

\subsection{Attention Mechanisms and Sparse Computation}

Attention has become the central computational primitive in modern vision networks, yet its quadratic cost with respect to sequence length remains a persistent bottleneck. Local window attention~\cite{liu2024vmamba} confines each token to a bounded spatial neighbourhood, improving throughput but weakening the long-range dependency modelling critical for small object detection. Deformable attention~\cite{zhu2021deformable} maintains linear cost by sampling a fixed number of reference points per query, but loses the ability to adapt to local feature complexity. Token merging approaches~\cite{bolya2023tome} combine redundant tokens to reduce sequence length, yet the resulting loss of positional specificity is detrimental for localisation-sensitive tasks. \rv{In detection, sparse query designs appear in Sparse 
DETR~\cite{roh2022sparse} and Efficient DETR~\cite{yao2021efficient}, which reduce computational load by selecting informative encoder tokens. However, these methods fix the sparsification threshold globally, whereas the informational density of a feature map varies substantially between defect-heavy industrial images and sparsely populated aerial frames. DCFA addresses this by predicting the retention ratio from local feature statistics, making sparsification content-dependent rather than scene-agnostic, and pairs the resulting sparse attention with a spatial gated linear unit that enriches nonlinear channel transformations with neighbourhood context.}

\subsection{Multi-Scale Feature Fusion}

Feature pyramids~\cite{lin2017fpn} established the 
principle of combining semantic richness from deep 
layers with spatial precision from shallow layers 
through a top-down pathway with lateral connections. 
PANet~\cite{liu2018panet} augmented this with a 
complementary bottom-up path, shortening the 
information routing distance between low-level detail 
and high-level semantics. BiFPN, introduced within 
EfficientDet~\cite{tan2020efficientdet}, went further 
by learning scalar weights for each cross-scale 
connection, allowing the network to modulate the 
relative contribution of each resolution level during 
fusion. \rv{Despite these refinements, a complication 
common to all pyramid designs has received little 
attention: upsampling expands feature map area by 
$s^2$, inflating the $L_1$ norm of activations and 
disrupting gradient balance between upsampled and 
lateral paths. SlimNeck~\cite{wang2024slimNeck} and 
Gold-YOLO~\cite{wang2023goldyolo} improve feature 
routing efficiency but leave this amplitude drift 
uncorrected. DFPN compensates through an analytically 
derived normalisation factor, providing stable 
cross-scale fusion that prior designs do not offer.}

\subsection{Frequency-Domain Representations in Neural Networks}

The motivation for incorporating spectral operations into detection networks
has both efficiency and representational grounds. Phase-aware MLP
architectures~\cite{tang2022wave} demonstrated that decomposing image patches
into amplitude and phase components enables richer spatial modelling than
purely spatial mixing, establishing that frequency-domain parameterisation
captures structure that spatial convolutions miss. Beyond representational
benefits, the computational argument is compelling: a convolution of kernel
size $k$ over a spatial map of size $N$ costs $\mathcal{O}(Nk^2)$, whereas
the equivalent spectral operation requires only $\mathcal{O}(N \log N)$
regardless of effective receptive field size.

Among classification-oriented frequency methods, \linebreak FcaNet~\cite{qin2021fca}
decomposes channel attention into discrete cosine components,
GFNet~\cite{rao2021gfnet} substitutes spatial token-mixing layers with global
Fourier filters, and FocalNet~\cite{yang2022focalnet} aggregates context at
multiple scales through focal modulation. These methods enrich feature
representations within a single resolution stage but do not engage with the
cross-scale fusion problem. \rv{Within the detection literature,
SF-DETR~\cite{wang2025sfdetr} injects scale-frequency awareness into the
encoder to capture high-frequency boundary cues in drone-view scenes, and
Freq-DETR~\cite{freqdetr2025} constructs dual-branch processing to jointly
model spatial and frequency features for UAV imagery. Both methods treat
frequency as a supplementary signal alongside spatial features, however,
rather than as the exclusive medium for bottleneck computation. FIRC3 departs
from this pattern by fully relocating feature refinement into the spectral
domain and solving the resulting problem as a constrained least-squares
optimisation, giving the network direct and learnable control over the
high-frequency content that governs boundary localisation for small objects.} 

\begin{figure*}[!t]
    \centering
    \includegraphics[width=0.95\linewidth]{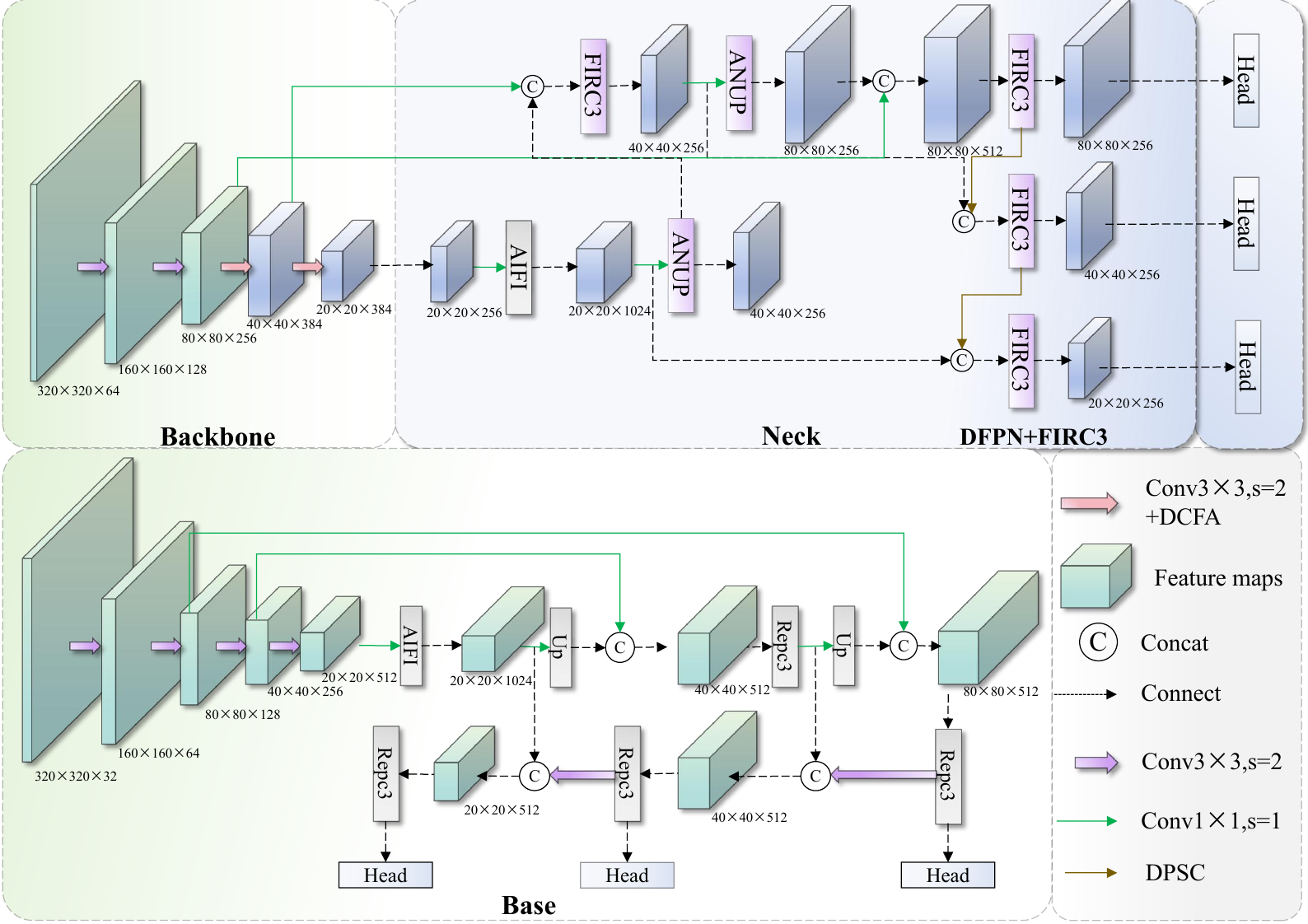}
    \caption{Overall architecture of DFIR-DETR.}
    \label{fig:network_architecture}
\end{figure*}

\section{Methods}
\label{sec:methods}

We propose DFIR-DETR to address small object detection challenges in UAV remote sensing and industrial surface inspection. The architecture, illustrated in Fig.~\ref{fig:network_architecture}, tackles two fundamental problems: sparse feature representation of distant small objects in VisDrone and subtle defect texture capture in NEU-DET. DCFA serves as the Backbone, where content-adaptive K-sparse attention allocates computational resources dynamically—defect regions and small targets receive concentrated modeling while uniform backgrounds undergo aggressive pruning. DFPN functions as the Neck, preventing feature inflation during upsampling through amplitude normalization and retaining fine spatial details through dual-path shuffle convolution across scales. FIRC3 operates in the feature fusion layer, processing features in the frequency domain to capture global context at reduced computational cost. Each component targets specific failure modes of existing detectors: DCFA addresses limited receptive fields, DFPN prevents information loss during scale transitions, and FIRC3 preserves high-frequency boundary details.

\subsection{DCFA}

\begin{figure*}[htbp]
    \centering
    \begin{subfigure}[t]{0.48\linewidth}
        \centering
        \includegraphics[width=0.94\linewidth]{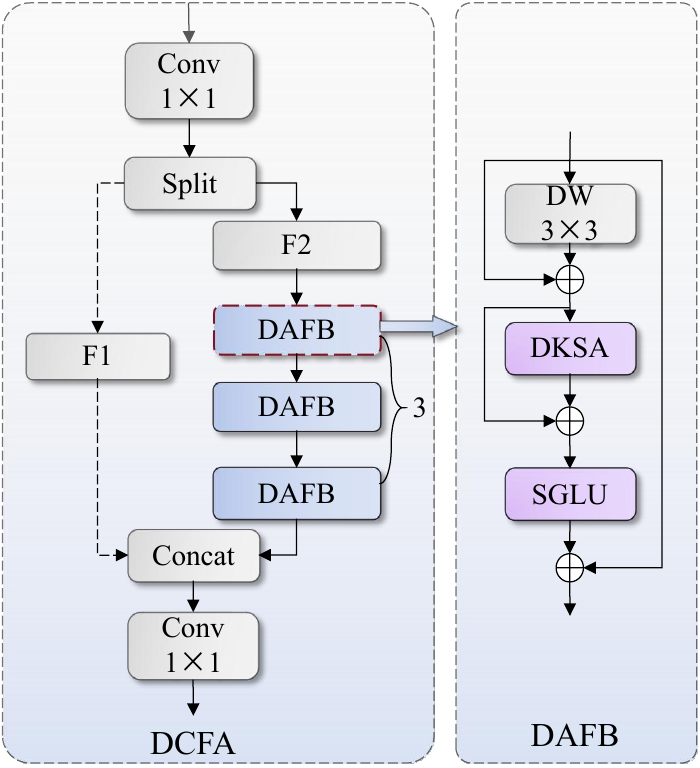}
        \caption{DCFA block}
        \label{fig:DCFA}
    \end{subfigure}
    \hfill
    \begin{subfigure}[t]{0.48\linewidth}
        \centering
        \includegraphics[width=\linewidth]{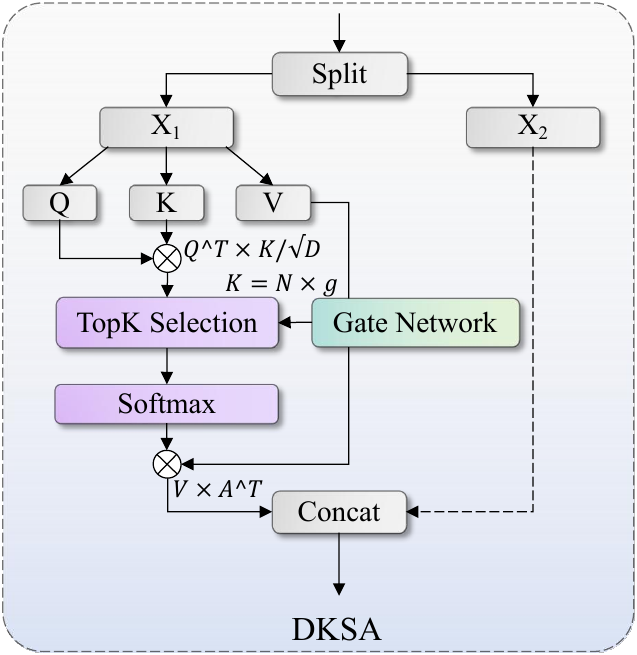}
        \caption{DKSA block}
        \label{fig:DKSA}
    \end{subfigure}
\end{figure*}

The backbone of conventional RT-DETR relies on a ResNet architecture
composed of homogeneous BasicBlock modules stacked uniformly across all
stages. However, this design exhibits three limitations that become
particularly pronounced in cross-scene small object detection. Progressive
downsampling erodes the fine-grained geometric detail and edge continuity
that small objects and subtle surface defects depend upon, while the fixed
receptive field expansion schedule lacks the flexibility to accommodate
the wide scale range encountered across different scenes. Furthermore,
without explicit spatial attention, the network cannot suppress irrelevant
background activations in dense or cluttered environments, leaving both
false positive and false negative rates persistently high. To address
these limitations, we propose DCFA, a content-adaptive feature aggregation
backbone that treats representational allocation as a learnable function
of local feature complexity. The key innovation is learning to allocate computational resources based on local complexity: small object regions receive concentrated attention modeling through dynamic K-sparse selection (DKSA), while uniform backgrounds are efficiently pruned. Spatial gated linear units (SGLU) further incorporate neighborhood context into nonlinear transformations. This design simultaneously preserves fine-grained geometric details and high-level semantic information, as illustrated in Fig.~\ref{fig:DCFA}.

\FloatBarrier

The DCFA module adopts a cross-stage partial (CSP) network architecture, achieving gradient-efficient propagation and multi-scale feature fusion through dual-path feature flows. The complete feature transformation process can be mathematically formulated as:

\begin{equation}
    \begin{aligned}
    Y = \phi_{cv2}\Big( \text{Concat}\big[ &F_{1},F_{2},G_{1}\left( F_{2} \right),G_{2}\left( G_{1}\left( F_{2} \right) \right),\\
    &\ldots,G_{n}\left( G_{n - 1}(\cdots) \right) \big] \Big)
    \end{aligned}
\end{equation}

\begin{equation}
[F_{1},\ F_{2}] = \text{split}\ (\phi_{cv1}\ (X))
\end{equation}

where $G_{i}$ represents the $i$-th DAFB (Dynamic Attention Fusion Block) transformation operator, $n$ denotes the stacking depth, and $\phi_{cv1}$ and $\phi_{cv2}$ represent the input and output $1 \times 1$ convolutional projection layers, respectively.

DAFB achieves comprehensive feature enhancement by cascading three functionally complementary sub-modules. Its forward propagation follows the transformation chain:

\begin{equation}
    \begin{split}
        Z &= \text{SGLU}\left( H + \text{DKSA}(H) \right), \\
        H &= X + \phi_{dw}(X)
    \end{split}
\end{equation}

where $\phi_{dw}$ denotes a $3 \times 3$ depthwise separable convolution with batch normalization. The DKSA mechanism subsequently operates on the enhanced features $H$, selectively focusing on defect regions in industrial scenarios through dynamic sparsification strategies while establishing long-range associations between small objects and contexts in remote sensing scenarios. SGLU serves as a feedforward network to perform nonlinear transformation and channel mixing on attention-enhanced features. Its gating mechanism adaptively modulates the activation intensity of different channels, further improving cross-scene feature discriminability. The mathematical expression is:

\begin{equation}
    \begin{aligned}
    \text{SGLU}(X) = X &+ \phi_{out}\Big( \mathcal{D}\big( \big( \phi_{act}\left( \phi_{dw}^{3 \times 3}\left( X_{g} \right) \right) \big) \\
    &\quad\odot X_{v} \big) \Big)
    \end{aligned}
\end{equation}

$\phi_{dw}^{3 \times 3}$ represents the $3 \times 3$ depthwise convolution applied to the gating stream, encoding local spatial contextual information into gating signals so that gating decisions integrate neighborhood spatial patterns rather than relying solely on single-point feature values. $\phi_{act}$ employs the GELU activation function, providing smoother gradient flow compared to ReLU. Element-wise multiplication $\odot$ implements gating modulation, and $\mathcal{D}$ denotes Dropout regularization.

DKSA implements computationally efficient self-attention through a dynamic Top-K selection strategy, demonstrating excellent adaptability in both industrial defect detection and remote sensing small object recognition scenarios, as illustrated in Fig.~\ref{fig:DKSA}. Combined with LGN (Layer Group Normalization) preprocessing, the complete attention computation process can be uniformly expressed as:

\begin{equation}
\text{DKSA}(X) = \phi_{proj}\left( \text{Concat}\left[ \left( VA^{T} \right)^{reshape},X_{2} \right] \right)
\end{equation}

\begin{equation}
A_{ij} = \left\{ \begin{matrix}
\frac{\exp\left( s_{ij} \right)}{\sum_{j' \in \mathcal{T}_{K}^{i}}\exp\left( s_{ij'} \right)}, & j \in \mathcal{T}_{K}^{i} \\
0, & j \notin \mathcal{T}_{K}^{i}
\end{matrix} \right.
\end{equation}

The dynamic Top-K selection mechanism is defined as:

\begin{equation}
K = \lfloor N \cdot \sigma\left( \text{AvgPool}\left( \psi(X) \right) \right)\rfloor
\end{equation}

where $\psi$ represents a gating network composed of two convolutional layers, $\sigma$ denotes the sigmoid function, and the output scalar value determines the proportion of attention connections to retain. $\mathcal{T}_{K}^{i} = \text{TopK}\left( S_{i},K \right)$ returns the index set of the top $K$ most relevant key positions for the $i$-th query position. Unselected connections are masked to zero, thereby converting the attention matrix from dense to structurally sparse, reducing computational complexity from $\mathcal{O}\left( N^{2} \right)$ to $\mathcal{O}(NK)$.

\FloatBarrier
\begin{algorithm}[H]
\small
\caption{DCFA}
\label{alg:dcfa}
\begin{algorithmic}[1]
    \REQUIRE Input feature map $X \in \mathbb{R}^{C \times H \times W}$, stack depth $N$
    \ENSURE Enhanced feature map $Y \in \mathbb{R}^{C \times H \times W}$
    
    \STATE $X_1, X_2 \leftarrow \text{Split}(X)$ \textit{// Dual-path feature flow}
    \STATE $F \leftarrow X_1$
    \FOR{$i = 1$ \textbf{to} $N$}
        \STATE \textit{// Dynamic Attention Fusion Block}
        \STATE $F' \leftarrow \text{DWConv}_{3\times3}(F) + \text{BN}(F)$
        
        \STATE \textit{// Dynamic K-Sparse Attention}
        \STATE $F'' \leftarrow \text{LGN}(F')$
        \STATE $Q, K, V \leftarrow \text{Linear}(F'')$
        \STATE $\rho \leftarrow \sigma(\text{Gate}(F''))$
        \STATE $k \leftarrow \lfloor \rho \cdot HW \rfloor$
        
        \FOR{\textbf{each} position $j$ \textbf{in} $[1, HW]$}
            \STATE $\mathcal{I}_k^j \leftarrow \text{TopK}(Q_j \cdot K^T, k)$
            \STATE $\text{Attn}_j \leftarrow \text{Softmax}\left(\frac{Q_j \cdot K_{\mathcal{I}_k^j}^T}{\sqrt{d}}\right) \cdot V_{\mathcal{I}_k^j}$
        \ENDFOR
        
        \STATE \textit{// Spatial Gated Linear Unit}
        \STATE $G \leftarrow \text{GELU}(\text{DWConv}_{3\times3}(\text{Attn}))$
        \STATE $F \leftarrow \text{Dropout}(\text{Linear}(\text{Attn}) \odot G)$
    \ENDFOR
    \STATE $Y \leftarrow \text{Conv}_{1\times1}(\text{Concat}(F, X_2))$
    \RETURN $Y$
\end{algorithmic}
\end{algorithm}

The dynamic K-sparse attention pruning mechanism enables content-adaptive allocation of computational resources, focusing more attention connections on defect regions in industrial scenarios to capture subtle textures while performing aggressive pruning on uniform backgrounds to reduce redundancy. In remote sensing scenarios, it allocates sufficient contextual modeling capability to small objects while sparsifying large irrelevant backgrounds, achieving optimal trade-offs between accuracy and efficiency in both scenarios. The spatial gated linear unit provides nonlinear modeling capability far exceeding traditional ReLU through spatialized gating signals. Its smooth gradient characteristics avoid gradient vanishing problems, enabling stable training of deep networks and learning of more refined feature transformations. The learnable gating mechanism in Eq.~(7) adaptively determines $K$ based on local feature statistics, which can be interpreted as adaptively balancing representational capacity against computational budget. This enables the network to automatically adjust the trade-off between efficiency and accuracy for each input sample based on scene complexity.

\subsection{DFPN}

The cross-scale feature fusion module CCFF in conventional RT-DETR relies on simple nearest neighbor upsampling and direct concatenation operations for multi-scale feature integration. However, abrupt transitions between feature scales introduce semantic discontinuities, compromising feature representation quality with particularly significant impact on small objects where spatial resolution is critical. The symmetric top-down pathway lacks sufficient consideration of information density differences between scales, resulting in suboptimal feature fusion performance in small object detection tasks. To address these fundamental deficiencies of CCFF, we propose DFPN, a principled redesign of multi-scale feature fusion guided by two core insights: upsampling must preserve feature intensity to prevent gradient instability, and downsampling must retain spatial details critical for small object boundaries. We achieve this through amplitude-normalized upsampling (ANUP) in the top-down pathway that maintains consistent feature magnitudes across scales, and dual-path shuffle convolution (DPSC) in the bottom-up pathway that explicitly preserves fine-grained spatial information during aggregation.

\sloppy
\rv{DFPN addresses feature inflation through \linebreak amplitude-normalised upsampling in the top-down pathway and recovers spatial detail via dual-path shuffle convolution in the bottom-up pathway, as illustrated in Fig.~\ref{fig:anup} and Fig.~\ref{fig:dpsc}.}
\fussy

\vspace{3mm}
\begin{center}
    \begin{minipage}[t]{0.48\linewidth}
        \centering
        \includegraphics[width=0.95\linewidth]{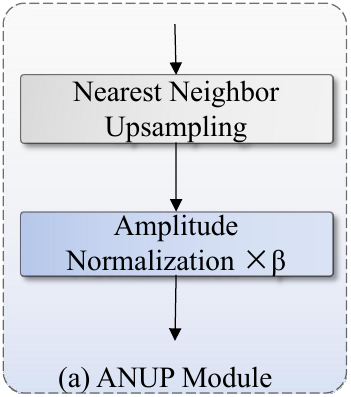}
        \captionof{figure}{ANUP module.}
        \label{fig:anup}
    \end{minipage}
    \hfill
    \begin{minipage}[t]{0.48\linewidth}
        \centering
        \includegraphics[width=1.05\linewidth]{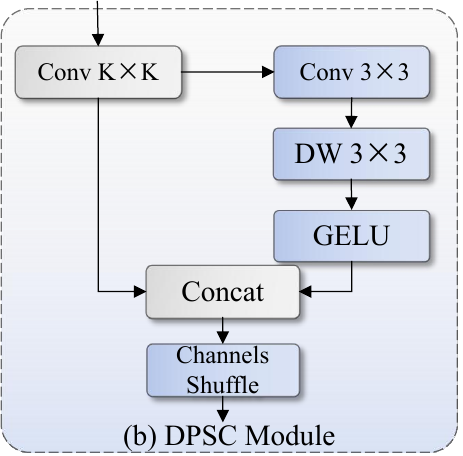}
        \captionof{figure}{DPSC module.}
        \label{fig:dpsc}
    \end{minipage}
\end{center}
\vspace{3mm}

In the top-down pathway, the ANUP module reconstructs traditional upsampling operations by introducing amplitude-dependent normalization factors to compensate for feature amplitude inflation caused by spatial expansion. Given an input feature map $F_{i}$ at scale $i$, the ANUP operation produces amplitude-controlled upsampled feature maps:

\begin{equation}
F_{i}^{\uparrow} = \beta \cdot \mathcal{U}\left( F_{i} \right) = \frac{1}{s^{2}} \cdot \mathcal{U}\left( F_{i} \right)
\end{equation}

where $\mathcal{U}( \cdot )$ denotes the nearest neighbor upsampling operator with scale factor $s$, and $\beta = 1/s^{2}$ represents the amplitude normalization coefficient, maintaining consistent feature intensity across different scales. This normalization strategy ensures that upsampled features retain original information density, preventing the feature map inflation common in traditional CCFF upsampling operations. The theoretical foundation stems from the fact that spatial interpolation inherently amplifies the feature map $L_{1}$ norm by a factor proportional to the square of the upsampling ratio. The proposed amplitude normalization counteracts this effect by maintaining the relationship:

\begin{equation}
    \begin{aligned}
    \| F_{i}^{\uparrow} \|_{1} &\approx \| F_{i} \|_{1} \cdot \frac{s^{2} \cdot H \cdot W}{H \cdot W} \cdot \frac{1}{s^{2}} \\
    &= \| F_{i} \|_{1}
    \end{aligned}
\end{equation}

establishing a more stable feature propagation mechanism, which is particularly significant for preserving subtle appearance features of small objects.

In the bottom-up pathway, the DPSC module addresses information loss during downsampling through a dual-path architecture. The first path performs standard convolution to extract semantic features, while the second path takes the output of the first path as input, capturing spatial details through cascaded convolutions, ultimately fusing dual-path information through channel shuffling. The DPSC operation is formulated as:

\begin{equation}
F_{out} = \mathcal{M}\Big( \text{Concat}\big( F_{1}^{std},\phi\left( W_{d}*\left( W_{conv}*F_{1}^{std} \right) \right) \big) \Big)
\end{equation}

 $F_{1}^{std}$ represents semantic features extracted through standard convolution with activation function $\sigma( \cdot )$, $W_{conv}$ and $W_{d}$ denote the $3\times 3$ standard convolution kernel and $3\times 3$ depthwise convolution kernel (groups=C/2) in the second path respectively, $\phi( \cdot )$ represents the GELU activation function providing smooth gradient flow, and $\mathcal{M}( \cdot )$ denotes the channel shuffle operation. The channel shuffle mechanism is mathematically represented as a tensor rearrangement operation:

\begin{equation}
    \begin{aligned}
    \mathcal{M}(F) = &\text{Reshape}\Big( \text{Permute}\big( \\
    &\quad\text{Reshape}\left( F,[B,2,C/2,H,W] \right),\\
    &\quad[0,2,1,3,4] \big),[B,C,H,W] \Big)
    \end{aligned}
\end{equation}

reorganizing feature channels to facilitate cross-path information exchange. The second path in DPSC constitutes a spatial detail capture branch, taking the output $F_{1}^{std}$ of the first path as input and achieving fine-grained feature extraction through cascaded convolution operations:

\begin{equation}
F_{2}^{dep} = \phi\left( W_{d}*\left( W_{conv}*F_{1}^{std} \right) \right)
\end{equation}

where $W_{conv}$ is a $3\times 3$ standard convolution kernel for initial spatial feature transformation, and $W_{d}$ is a $3\times 3$ depthwise convolution kernel for efficient spatial filtering. Through grouped operations, spatial features are progressively refined while maintaining computational efficiency, with GELU activation ensuring smooth gradient propagation required for training stability. The concatenation and shuffling of features from both paths establish comprehensive feature representations, integrating semantic information from the standard convolution path with fine-grained spatial details from the cascaded convolution path.

\begin{algorithm}[t]
\small
\caption{DFPN}
\label{alg:dfpn}
\begin{algorithmic}[1]
    \REQUIRE Multi-scale features $\{F_i\}_{i=1}^L$, scale $s$
    \ENSURE Enhanced features $\{F_i^{\text{out}}\}_{i=1}^L$
    
    \STATE $F_L^{\text{up}} \leftarrow F_L$
    \STATE \textbf{\textit{// Top-down with ANUP}}
    \FOR{$i = L-1$ \textbf{down to} $1$}
        \STATE $F_i^{\text{up}} \leftarrow \textsc{ANUP}(F_{i+1}^{\text{up}}, F_i, s)$
    \ENDFOR
    
    \STATE \textbf{\textit{// Apply DPSC}}
    \FOR{$i = 1$ \textbf{to} $L$}
        \STATE $F_i^{\text{out}} \leftarrow \textsc{DPSC}(F_i^{\text{up}})$
    \ENDFOR
    
    \vspace{1em}
    \STATE \textbf{Procedure \textsc{ANUP}}$(F_{\text{high}}, F_{\text{low}}, s)$:
    \STATE \quad $F_{\text{up}} \leftarrow \text{Upsample}(F_{\text{high}}, s)$
    \STATE \quad $F_{\text{norm}} \leftarrow (1/s^2) \cdot F_{\text{up}}$ \textit{// Amplitude normalization}
    \STATE \quad \textbf{return} $\text{Conv}_{1\times1}(\text{Concat}(F_{\text{norm}}, F_{\text{low}}))$
    
    \vspace{1em}
    \STATE \textbf{Procedure \textsc{DPSC}}$(F)$:
    \STATE $F_1 \leftarrow \sigma(W_{\text{std}} \ast F)$ \textit{// Path 1}
    \STATE \quad $F_2 \leftarrow \text{GELU}(W_d \ast \text{GELU}(W_{\text{pw}} \ast F))$ \textit{// Path 2}
    \STATE \quad \textbf{return} $\text{ChannelShuffle}(\text{Concat}(F_1, F_2))$
\end{algorithmic}
\end{algorithm}
\FloatBarrier

ANUP stabilises top-down propagation by eliminating the norm inflation
inherent in standard upsampling, ensuring high-level semantic information
reaches lower scales with preserved intensity, a property particularly
valuable for small objects that depend heavily on contextual cues. DPSC
complements this by routing fine-grained boundary detail through a dedicated
cascaded convolution stream, so spatial precision is not sacrificed during
bottom-up aggregation. The result is more consistent cross-scale feature
flow and improved localisation accuracy for small objects across both
industrial and aerial detection scenarios.

\subsection{FIRC3}

\begin{figure*}[!t]
    \centering
    \includegraphics[width=0.8\linewidth]{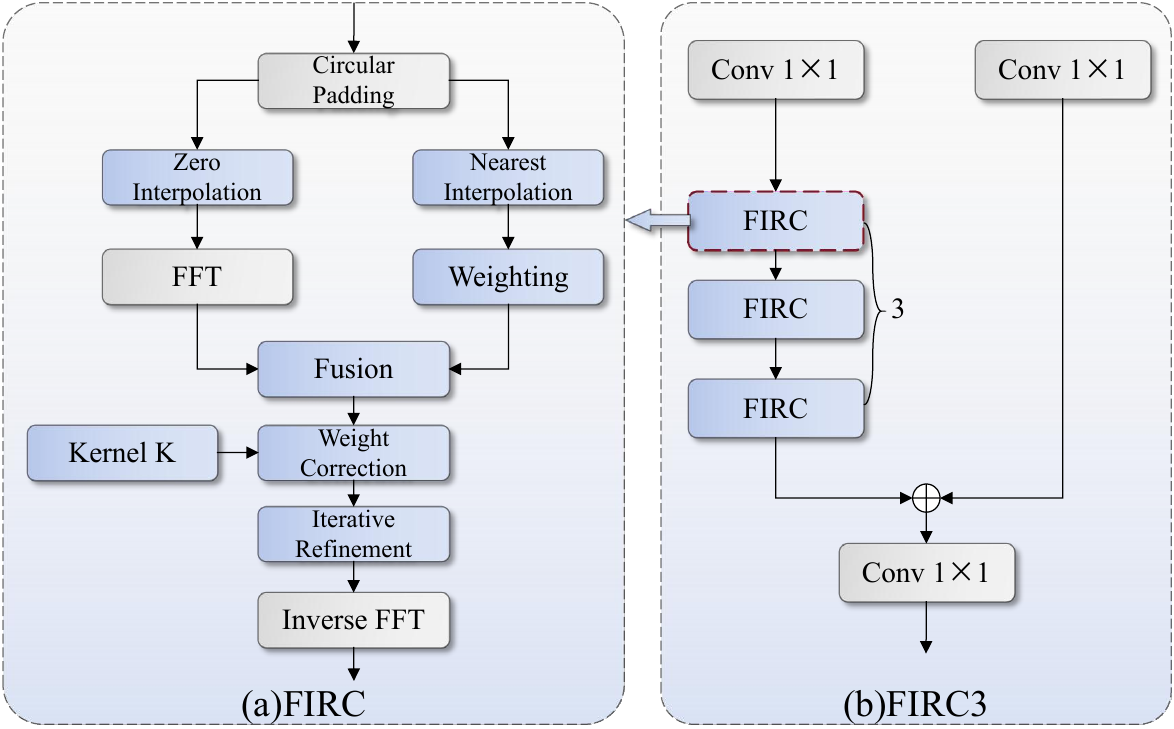}
    \caption{FIRC3 block}
    \label{fig:FIRC3}
\end{figure*}

Conventional RT-DETR employs the RepC3 module for multi-scale feature fusion in its feature pyramid network and path aggregation network structures. However, RepC3 primarily relies on spatial domain convolution operations, exhibiting inherent limitations in capturing long-range dependencies and processing small object features. The spatial locality constraint of conventional convolutions limits the receptive field scope, resulting in insufficient feature representation capability for small objects, while contextual information is crucial for small object detection. Repeated application of spatial domain convolutions introduces cumulative information loss, particularly severely affecting fine-grained details that small object detection depends upon. Targeting these limitations, we propose a novel feature aggregation module FIRC3, which leverages frequency domain transformations to enhance feature propagation and maintain the integrity of small object information throughout the detection pipeline, as illustrated in Fig.~\ref{fig:FIRC3}.

\FloatBarrier

The input feature map $X$ first generates intermediate representations through two parallel $1 \times 1$ convolutions. One branch undergoes cascaded FIRC (Frequency Iterative Refinement Convolution) transformation processing, while the other \linebreak branch maintains the original feature flow. The module output is formulated as:

\begin{equation}
Y = W_{3}\left( \mathcal{M}\left( W_{1}*X \right) + W_{2}*X \right)
\end{equation}

where $W_{1},W_{2}$ represent channel projection operators, $\mathcal{M}( \cdot )$ denotes the transformation sequence composed of $n$ cascaded FIRC operations, $W_{3} $ performs final channel mapping, $*$ denotes convolution operation, and the hidden channel dimension $C' = \lfloor e \cdot C_{\text{out}}\rfloor$ is controlled by expansion ratio $e$. The FIRC operator executes feature transformations in the frequency domain through a convex optimization framework. For a given feature map $F$, the FIRC operation first applies circular padding to enable seamless frequency domain processing, subsequently generating sparse and dense upsampled features through zero-interpolation upsampling $\mathcal{U}_{s}(F)$ and nearest-neighbor interpolation $\mathcal{I}_{s}(F)$ respectively. Define the frequency domain intermediate variable:

\begin{equation}
    \begin{aligned}
    F_{R} &= \mathcal{F}[K]^{*} \odot \mathcal{F}\left[ \mathcal{U}_{s}(F) \right] \\
    &\quad+ \mathcal{F}\left[ \epsilon_{b} \cdot \mathcal{I}_{s}(F) \right]
    \end{aligned}
\end{equation}

where $\mathcal{F}[ \cdot ]$ denotes the Fast Fourier Transform, $K$ represents the depthwise separable frequency domain convolution kernel initialized through softmax normalization, $( \cdot )^{*}$ denotes complex conjugate operation, $\odot$ represents element-wise product, $\epsilon_{b} = \sigma(b - 9.0) + \epsilon$ is the adaptive regularization term, $\sigma( \cdot )$ is the sigmoid function, and $\epsilon = 10^{-5}$ ensures numerical stability. This formula decomposes the original features into two pathways: the zero-interpolation path preserves precise information at original sampling positions, while the dense interpolation path provides smooth spatial continuity. Through the conjugate of the frequency domain convolution kernel and linear combination of both paths, adaptive fusion of multi-scale features is achieved. Based on $F_{R}$, the adaptive weight correction factor is computed:

\begin{equation}
W_{\text{inv}} = \frac{\text{Avg}_{s}\left( \mathcal{F}[K] \odot F_{R} \right)}{\text{Avg}_{s}\left( \left| \mathcal{F}[K] \right|^{2} \right) + \epsilon_{b}}
\end{equation}

The averaging operator $\text{Avg}_{s}( \cdot )$ performs spatial downsampling on frequency domain responses, implementing \linebreak frequency-selective aggregation by rearranging the spectrum into $s \times s$ blocks and averaging along the last dimension. The numerator term $\mathcal{F}[K] \odot F_{R}$ computes the response intensity between the frequency domain convolution kernel and intermediate features, while the denominator term $\left| \mathcal{F}[K] \right|^{2}$ represents the squared magnitude of the frequency domain energy spectrum, serving a normalization role to prevent excessive amplification of frequency domain responses. The final output is obtained through frequency domain iterative solving:

\begin{equation}
    \begin{aligned}
    \widehat{F}_{\text{out}} &= \mathcal{F}^{-1}\left[ \frac{F_{R} - \mathcal{F}[K]^{*} \odot \text{Repeat}_{s}\left( W_{\text{inv}} \right)}{\epsilon_{b}} \right]
    \end{aligned}
\end{equation}

where $\mathcal{F}^{-1}[ \cdot ]$ denotes the inverse Fourier transform, and $\text{Repeat}_{s}( \cdot )$ upsamples the downsampled weight factor to the original frequency domain resolution. The subtracted term $\mathcal{F}[K]^{*} \odot \text{Repeat}_{s}\left( W_{\text{inv}} \right)$ in the numerator can be understood as refined correction of coarse-scale responses, implementing backpropagation through frequency domain conjugate operations, thereby suppressing low-frequency redundancy while maintaining high-frequency details. The entire transformation process essentially solves a \linebreak frequency-domain-constrained least squares problem, adaptively balancing contributions of different frequency components and enabling the network to dynamically adjust sensitivity to high-frequency information of small objects. Periodization processing of the frequency domain convolution kernel is achieved through point spread function to optical transfer function conversion:

\begin{equation}
    \begin{aligned}
    \mathcal{F}\left[ K_{\text{circ}} \right] &= \mathcal{F}\Big[ \text{Roll}\big( \text{ZeroPad}\left( K,\left( H_{s},W_{s} \right) \right),\\
    &\quad\left( -\lfloor k/2\rfloor, -\lfloor k/2\rfloor \right) \big) \Big]
    \end{aligned}
\end{equation}

where $\text{ZeroPad}( \cdot )$ pads the convolution kernel to the target feature map size, and $\text{Roll}( \cdot )$ performs circular shift to align the kernel center to the coordinate origin, thereby achieving true circular convolution. Frequency domain operations possess the natural advantage of global receptive fields compared to spatial domain convolutions, enabling the network to capture long-distance feature dependencies without increasing parameter count. Through learnable frequency domain convolution kernel parameterization, the module can dynamically adjust sensitivity to different frequency components, better adapting to feature distribution characteristics of objects at different scales.

\FloatBarrier
\begin{algorithm}[!t]
\caption{FIRC3}
\label{alg:firc3}
\begin{algorithmic}[1]
\REQUIRE Input feature $X \in \mathbb{R}^{C \times H \times W}$, iterations $T$
\ENSURE Refined feature $Y \in \mathbb{R}^{C \times H \times W}$

\STATE $X_1, X_2 \leftarrow \text{Split}(X)$ \textit{// Dual-path split}
\STATE $F \leftarrow X_1$
\FOR{$t = 1$ \textbf{to} $T$}
    \STATE \textit{// Frequency-domain transformation}
    \STATE $F_{\text{freq}} \leftarrow \text{FFT}(F)$
    \STATE $F_{\text{amp}} \leftarrow |F_{\text{freq}}|$ \textit{// Amplitude}
    \STATE $F_{\text{phase}} \leftarrow \angle F_{\text{freq}}$ \textit{// Phase}
    
    \STATE \textit{// Frequency-domain feature processing}
    \STATE $F_{\text{amp}}' \leftarrow \text{Conv}_{1\times1}(F_{\text{amp}})$
    \STATE $F_{\text{phase}}' \leftarrow \text{Conv}_{1\times1}(F_{\text{phase}})$
    
    \STATE \textit{// Reconstruct and inverse transform}
    \STATE $F_{\text{freq}}' \leftarrow F_{\text{amp}}' \cdot e^{i \cdot F_{\text{phase}}'}$
    \STATE $F' \leftarrow \text{IFFT}(F_{\text{freq}}')$
    
    \STATE \textit{// Residual connection}
    \STATE $F \leftarrow F + \text{GELU}(\text{Conv}_{3\times3}(F'))$
\ENDFOR
\STATE $Y \leftarrow \text{Conv}_{1\times1}(\text{Concat}(F, X_2))$
\RETURN $Y$
\end{algorithmic}
\end{algorithm}
\FloatBarrier

Operating in the spectral domain grants FIRC3 an implicit global receptive
field at $\mathcal{O}(N\log N)$ cost, while the iterative correction
mechanism suppresses low-frequency redundancy and restores the
boundary-defining high-frequency signal that repeated spatial convolutions
progressively attenuate. The result is more precise localisation of
small objects whose appearance is predominantly encoded in fine-scale
edge structure. 
\section{Experiments}
\label{sec:experiments}

\subsection{Experimental Setup}

All experiments were conducted on an NVIDIA GeForce RTX~3090 GPU.
DFIR-DETR was implemented in\linebreak PyTorch~2.2.2 / TorchVision~0.17.2, configured
based on RT-DETR-R18. Training used AdamW with learning rate $1\times10^{-4}$
and weight decay $5\times10^{-4}$, for 300 epochs with batch size 4, input
resolution $640\times640$, and mixed-precision training. Average GPU memory
consumption was 11 GB and training required 7 hours per
dataset.

\subsection{Benchmark Datasets}

\rv{These two benchmarks were not chosen arbitrarily. Prior work on small object detection tends to evaluate on a single domain, leaving open the question of whether reported gains reflect architectural improvements or dataset-specific overfitting. VisDrone and NEU-DET differ in sensor type, scene structure, object formation mechanism, and background statistics, but they both expose the same underlying bottleneck: insufficient high-frequency boundary representation for objects with limited pixel support. Demonstrating consistent gains across both settings provides stronger evidence that the proposed modules address a general architectural deficiency rather than a narrow application-specific one.}

\begin{figure*}[!t]
    \centering
    \includegraphics[width=0.83\textwidth]{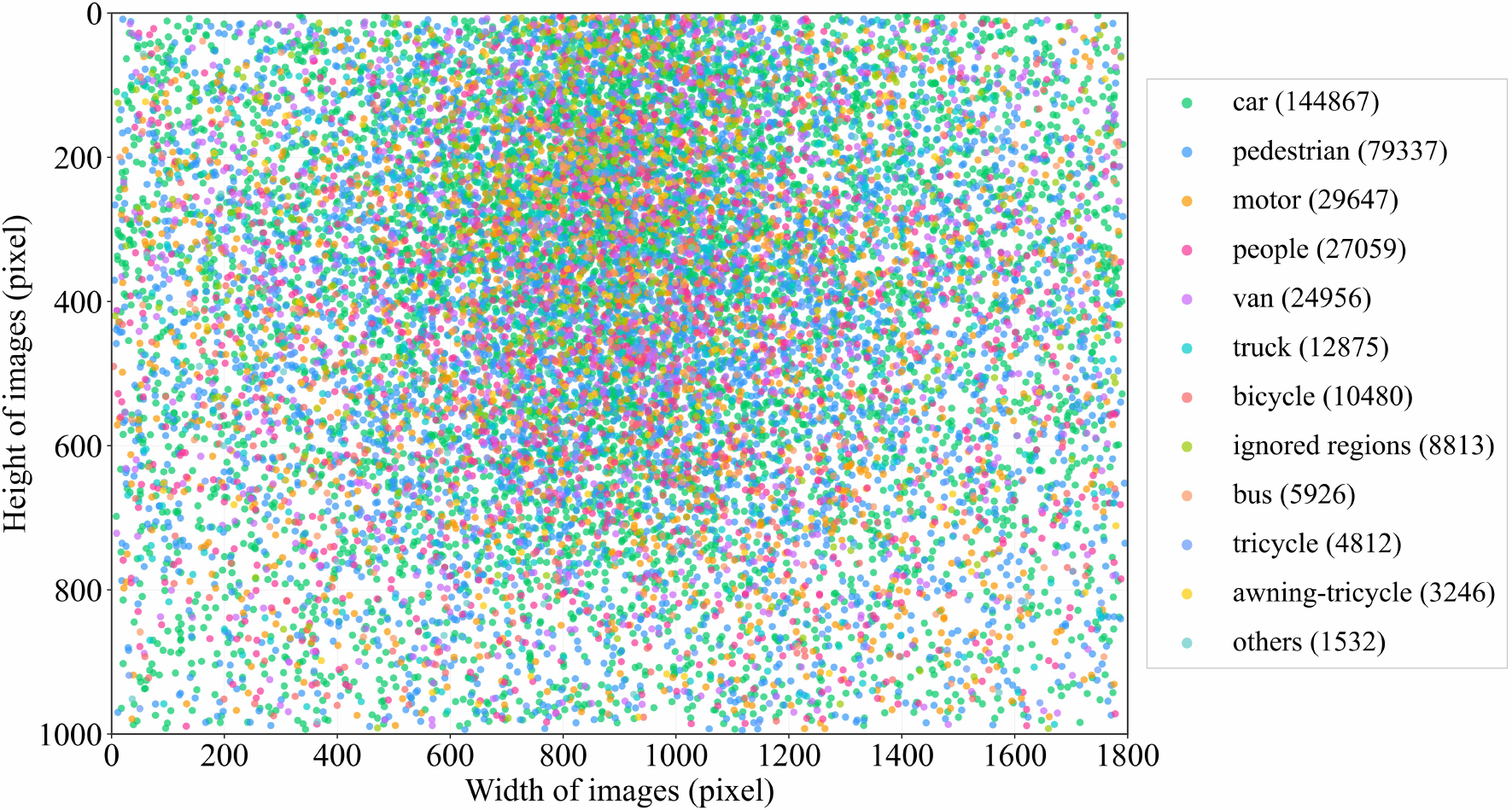}
    \caption{VisDrone object instance spatial distribution.}
    \label{fig:visdrone_distribution}
\end{figure*}

\textbf{VisDrone}~\cite{zhu2021visdrone} contains 10,209 aerial images with
2.6M bounding boxes across 10 categories. The dataset splits into 6,471
training, 548 validation, and 3,190 testing images, presenting challenges of
large scale variations, dense occlusions, and viewpoint diversity, as illustrated in Fig.~\ref{fig:visdrone_distribution}.

\begin{figure*}[!t]
    \centering
    \includegraphics[width=0.83\textwidth]{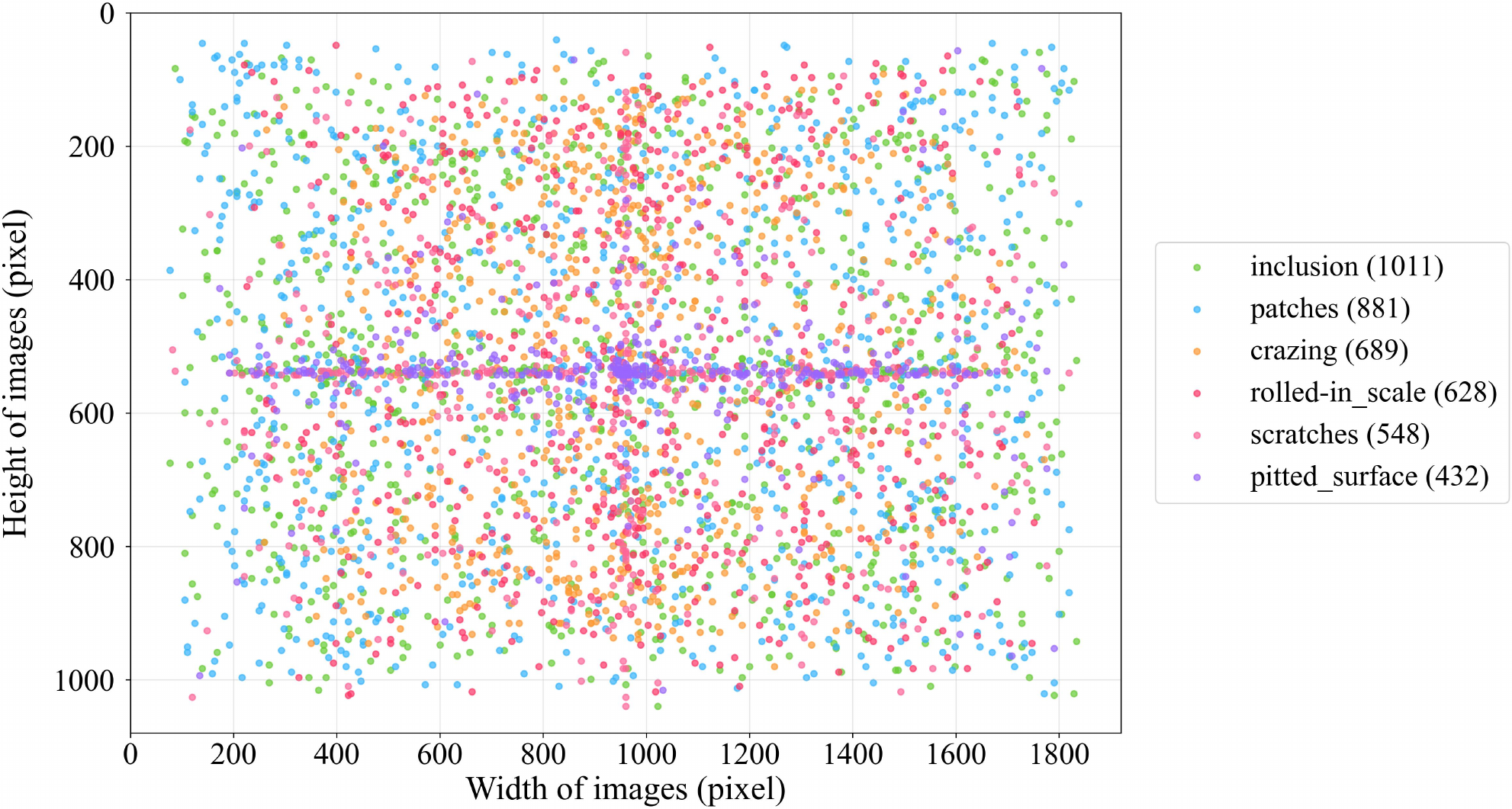}
    \caption{NEU-DET defect instance spatial distribution.}
    \label{fig:neudet_distribution}
\end{figure*}

\textbf{NEU-DET}~\cite{song2013neu} comprises 1,800 grayscale images of
hot-rolled steel surfaces with six defect types (crazing, inclusion, patches,
pitted surface, rolled-in scale, scratches), each with 300 images at
$200\times200$ pixels. Challenges include subtle texture variations, complex
backgrounds, and high inter-class similarity, as illustrated in Fig.~\ref{fig:neudet_distribution}.

\subsection{Ablation Study}

Table~\ref{tab:ablation_neudet} reports ablation results on NEU-DET 
under two complementary protocols: sequential module addition and 
single-module removal from the complete model. DCFA alone lifts mAP50 
by 1.4 points over the RT-DETR baseline while trimming parameters from 
19.9M to 13.7M; the 31.2\% reduction comes from the sparse attention 
structure, not from any sacrifice in representational depth. Substituting 
only DFPN into the baseline reaches 90.9\% mAP50, with the Ri category 
gaining 11.2 points over baseline, the largest single-category swing 
in the table; this is consistent with ANUP counteracting the norm 
inflation that standard upsampling introduces at that scale. FIRC3 
alone reaches 90.5\% mAP50 at the lowest GFLOPs of any single-module 
configuration, reflecting the computational advantage of spectral 
operations over repeated spatial convolutions. The complete model 
reaches 92.9\% mAP50 and 65.9\% mAP50:95; the 7.7-point mAP50:95 
margin over baseline outpaces the mAP50 margin, which points to 
boundary localisation rather than coarse classification as the 
primary beneficiary. Parameter count and GFLOPs land at 11.7M and 
47.2, down 41.2\% and 17.2\% from the starting point.

\rv{Removal experiments invert the construction protocol. Taking DCFA 
out of the full model costs 1.8 mAP50 points; removing DFPN costs 
1.2 points; removing FIRC3 costs 0.5 points. The ordering matches 
the sequential results: DCFA contributes most on NEU-DET, where 
content-adaptive sparsification has the sharpest effect on defect 
boundary \linebreak discrimination, and confirms that the gains are not an 
artefact of module interaction alone.}

\begin{table*}[!t]
    \renewcommand{\arraystretch}{1.3}
    \footnotesize
    \caption{\centering Ablation study on NEU-DET dataset.}
    \label{tab:ablation_neudet}
    \centering
    \setlength{\tabcolsep}{3.5pt}
    \begin{tabular}{cccccccccccccc}
        \toprule
        DCFA & DFPN & FIRC3 & mAP50 & Cr & In & Pa & Ps & Ri & Sr
             & mAP50:95 & Params & GFLOPs & FPS \\
        \midrule
        $\times$ & $\times$ & $\times$
            & 88.7 & 78.4 & 87.0 & 96.0 & 94.6 & 78.0 & 98.1
            & 58.2 & 19.9 & 57.0 & 36.9 \\
        \checkmark & $\times$ & $\times$
            & 90.1 & 78.9 & 91.2 & \textbf{97.0} & 96.8 & 84.0 & 98.6
            & 60.8 & 13.7 & 51.7 & 37.0 \\
        \rv{$\times$} & \rv{\checkmark} & \rv{$\times$}
            & \rv{90.9} & \rv{72.1} & \rv{94.5} & \rv{97.0} & \rv{94.4}
            & \rv{89.2} & \rv{98.3}
            & \rv{62.6} & \rv{19.7} & \rv{56.7} & \rv{33.7} \\
        \rv{$\times$} & \rv{$\times$} & \rv{\checkmark}
            & \rv{90.5} & \rv{76.2} & \rv{94.6} & \rv{95.8} & \rv{92.8}
            & \rv{84.7} & \rv{98.6}
            & \rv{63.2} & \rv{18.2} & \rv{48.4} & \rv{47.8} \\
        \checkmark & \checkmark & $\times$
            & 92.4 & \textbf{86.1} & 92.4 & 96.6 & \textbf{98.8} & 84.8 & 95.6
            & 64.0 & 13.4 & 49.8 & \textbf{40.2} \\
        \rv{\checkmark} & \rv{$\times$} & \rv{\checkmark}
            & \rv{91.7} & \rv{82.9} & \rv{92.1} & \rv{94.8} & \rv{98.3}
            & \rv{83.2} & \rv{99.1}
            & \rv{62.1} & \rv{14.2} & \rv{48.2} & \rv{37.6} \\
        \rv{$\times$} & \rv{\checkmark} & \rv{\checkmark}
            & \rv{91.1} & \rv{81.3} & \rv{89.9} & \rv{96.7} & \rv{95.1}
            & \rv{84.7} & \rv{98.7}
            & \rv{61.6} & \rv{15.1} & \rv{47.3} & \rv{33.7} \\
        \checkmark & \checkmark & \checkmark
            & \textbf{92.9} & 84.0 & \textbf{93.7} & 96.8 & 98.4
            & \textbf{84.9} & \textbf{99.3}
            & \textbf{65.9} & \textbf{11.7} & \textbf{47.2} & 38.0 \\
        \bottomrule
    \end{tabular}
\end{table*}

\subsection{Comparative Performance Analysis and Qualitative Visualisation}

\rv{On NEU-DET, DFIR-DETR reaches 92.9\% mAP50, surpassing the RT-DETR 
baseline by 4.2 points and YOLOv11m by 0.3 points while reducing 
parameters to 11.7M, a 77.2\% reduction relative to YOLOv6m at 
substantially higher accuracy, as shown in Table~\ref{tab:comparison_neudet}. 
The mAP50:95 gap over baseline is 7.7 points, larger than the mAP50 gap, 
indicating that frequency-domain boundary recovery contributes more to 
tight localisation than to coarse detection. Beyond general-purpose 
detectors, Table~\ref{tab:comparison_neudet} also incorporates three methods 
developed specifically for steel surface inspection: \linebreak
IMN-YOLOv3~\cite{zhang2022imnyolo}, FC-DETR~\cite{li2024fcdetr}, and 
MDT-Net~\cite{wang2025mdtnet}. Despite their domain-specific design, the best-performing specialist 
method, IMN-YOLOv3, achieves 86.9\% mAP50, leaving a gap of at least 
6.0 points relative to DFIR-DETR's 92.9\%. Concurrently, FC-DETR requires 56.5M parameters to 
achieve 83.7\%, nearly five times the footprint of our model at comparable 
or lower accuracy. The pattern suggests that frequency-domain boundary 
modelling addresses a representational gap that domain-specific architectural 
choices \linebreak alone have not closed.}

\begin{figure*}[!t]
    \centering
    \includegraphics[width=0.95\textwidth]{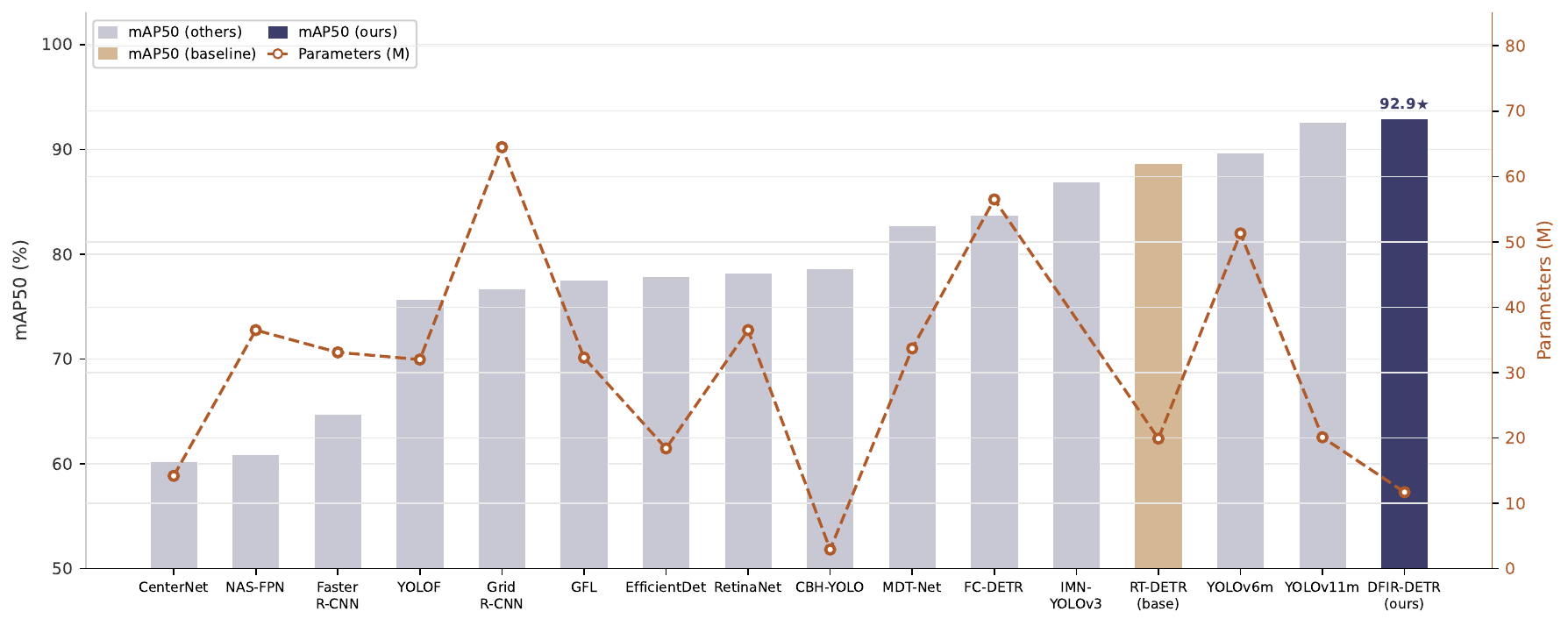}
    \caption{Comparison on NEU-DET.}
    \label{fig:radar_neudet}
\end{figure*}

At the component level, Table~\ref{tab:backbone_neudet} shows that DCFA 
achieves 90.1\% mAP50 on NEU-DET, outperforming ELGCA by 1.9 points despite 
comparable parameter counts, with particularly strong gains on crazing 
(+16.9 points over ELGCA). On VisDrone, Table~\ref{tab:backbone_visdrone} 
shows that DCFA reaches the highest mAP50:95 of 30.1\% among all backbone 
alternatives, suggesting that content-adaptive attention allocation provides 
more consistent localisation across the ten aerial categories than 
geometry-driven or token-reduction approaches. As reported in 
Table~\ref{tab:neck_neudet}, DFPN achieves 90.9\% mAP50 on NEU-DET, a 
3.1-point margin over MSGA, with the Ri category showing the largest 
per-class gain (+6.7 points), consistent with ANUP preserving the fine-scale 
edge structure that rolled-in scale defects require. 
Table~\ref{tab:repc3_neudet} shows that FIRC3 reaches 90.5\% mAP50, 
outperforming DGCST by 2.7 points, while also achieving the lowest GFLOPs 
among all compared fusion modules, as illustrated in 
Fig.~\ref{fig:radar_neudet}.

\begin{figure*}[!t]
    \centering
    \includegraphics[width=0.95\textwidth]{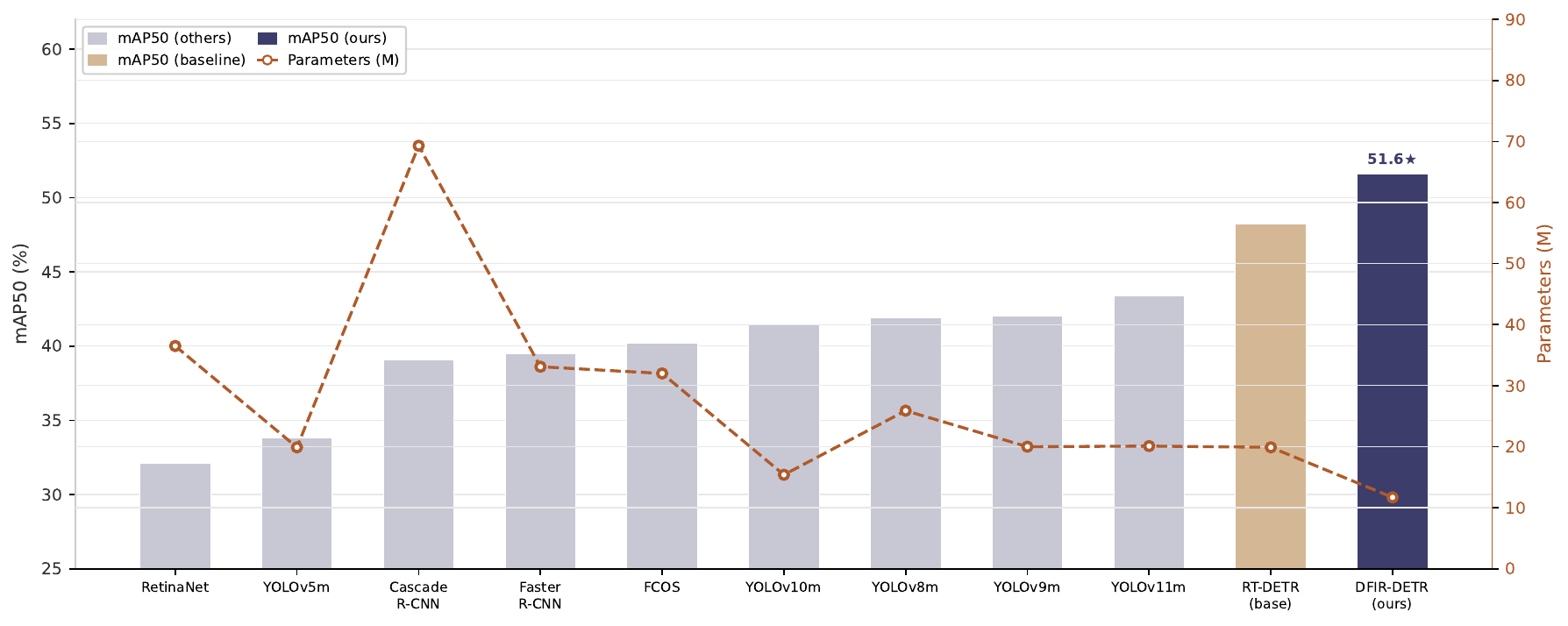}
    \caption{Comparison on VisDrone.}
    \label{fig:radar_visdrone}
\end{figure*}

On VisDrone, DFIR-DETR achieves 51.6\% mAP50, a 3.4-point improvement 
over the RT-DETR baseline and 8.2 points over YOLOv11m, as reported in 
Table~\ref{tab:comparison_visdrone}. The awning-tricycle category shows 
the largest per-class gain (+11.7 points over baseline), a category 
combining small physical size with an unusual aspect ratio where 
receptive-field adaptation and high-frequency boundary retention matter 
most, as illustrated in Fig.~\ref{fig:radar_visdrone}. \rv{Table~\ref{tab:comparison_visdrone_uav} further compares DFIR-DETR against UAV-specific detectors under the identical 640$\times$640 input protocol. DFIR-DETR surpasses MFP-YOLO and CRL-YOLOv5 by 5.6 and 12.4 percentage points in mAP50 respectively, while achieving the lowest parameter count among all compared Transformer-based methods.}

\begin{figure*}[!t]
\centering
\includegraphics[width=0.95\textwidth]{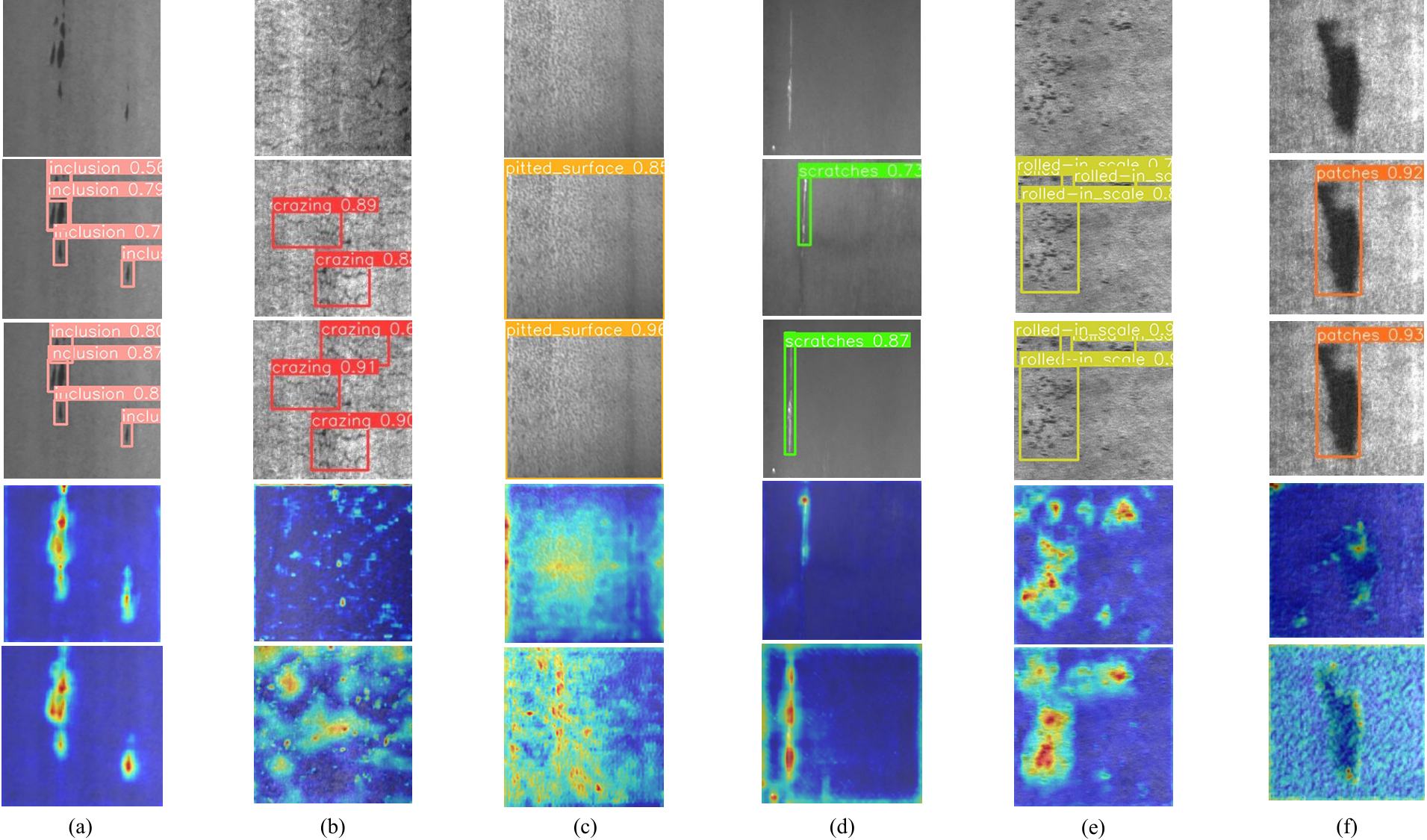}
\caption{Qualitative visualization comparison on NEU-DET dataset.}
\label{fig:visualization}
\end{figure*}

Fig.~\ref{fig:visualization} presents qualitative detection results and GradCAM activation maps on NEU-DET across all six defect categories. Each column corresponds to one category: (a)~inclusion, (b)~crazing, (c)~pitted surface, (d)~scratches, (e)~rolled-in scale, and (f)~patches. For each column, the top row shows the ground truth annotation, the second and third rows show detection results from the RT-DETR baseline and DFIR-DETR respectively, and the bottom two rows show the corresponding GradCAM activation maps.

Across all six categories, DFIR-DETR produces tighter bounding boxes at higher confidence than the baseline. For spatially elongated defects such as scratches (d) and rolled-in scale (e), the baseline tends to produce loose or fragmented boxes, whereas DFIR-DETR recovers more precise boundary extents. The GradCAM maps tell the same story from the feature side: baseline activations spread broadly across the feature map, whereas DFIR-DETR pulls high-activation regions onto defect boundaries. This contrast is most pronounced in the inclusion (a) and pitted surface (c) categories, where subtle low-contrast textures demand precise high frequency boundary discrimination. The tighter activation footprint is a direct visual signature of the spectral boundary reinforcement that FIRC3 introduces.
\section{Conclusion}
\label{sec:conclusion}
\balance

Small object detection in complex, cross-scene environments 
exposes fundamental tensions in neural network design that 
cannot be resolved by scaling existing architectures. This 
work argues that three specific structural properties (uniform 
attention allocation, amplitude drift during multi-scale 
fusion, and spectral attenuation through repeated spatial 
convolution) collectively account for a substantial portion 
of the performance gap between current detectors and the 
demands of real-world deployment. DFIR-DETR addresses each 
through principled architectural interventions: 
content-adaptive sparse attention in DCFA, norm-preserving 
feature pyramid construction in DFPN, and frequency-domain 
iterative refinement in FIRC3. The resulting framework 
demonstrates that targeted, theoretically motivated 
modifications to backbone, neck, and fusion modules can 
simultaneously improve accuracy, reduce parameters, and 
maintain real-time throughput without relying on larger 
training data or heavier model capacity.

\rv{Three concrete directions emerge from the limitations 
of the current design. The frequency-domain computation 
in FIRC3 stops at the encoder: decoder cross-attention 
operates entirely in the spatial domain, and there is no 
mechanism to bias query-key interactions toward 
boundary-informative spectral components. Introducing a 
spectral regularisation term into decoder attention could 
tighten localisation for the sub-16-pixel targets where 
spatial cues are weakest. The norm-preserving argument 
behind DFPN currently assumes \linebreak nearest-neighbour upsampling; 
the derivation does not carry over to learnable operators 
such as CARAFE or DySample, whose sampling grids introduce 
non-uniform amplitude effects that the present normalisation 
factor does not account for. Finally, DCFA commits to a 
single sparsity ratio for an entire feature map, which 
treats a defect cluster and an empty background patch as 
equally informative. At a modest spatial resolution, Patch-level ratio prediction could redistribute attention 
budget more precisely without the overhead of full dense 
computation.}

\begin{table*}[!t]
    \renewcommand{\arraystretch}{1.3}
    \footnotesize
    \caption{\centering Backbone comparison on VisDrone dataset.}
    \label{tab:backbone_visdrone}
    \centering
    \setlength{\tabcolsep}{2.8pt}
    \begin{tabular}{lccccccccccccccc}
    \toprule
    Model & mAP50 & Ped & Peo & Bic & Car & Van & Tru & Tri & Awn
          & Bus & Mot & mAP95 & Params & GFLOPs & FPS \\
    \midrule
    Strip~\cite{guo2023strip}
        &48.4&53.8&36.6&23.9&81.4&54.5&57.7&\textbf{35.7}&23.9&68.8&49.9
        &28.8&14.3&47.7&36.1\\
    MambaOut~\cite{yu2024mambaout}
        &46.7&51.1&34.0&22.9&80.2&51.8&53.9&34.4&24.4&65.9&48.2
        &27.8&15.9&\textbf{41.9}&34.7\\
    GlobalFilter~\cite{rao2022global}
        &48.4&53.6&36.7&25.5&81.4&53.4&55.7&34.1&27.4&66.3&50.0
        &28.7&16.9&46.5&36.9\\
    AP~\cite{dai2017ap}
        &48.6&53.9&35.8&23.8&81.8&54.6&56.6&34.8&\textbf{28.1}&67.4&49.8
        &28.9&13.3&47.0&36.9\\
    FDT~\cite{mao2023fdt}
        &47.2&51.7&34.5&24.0&81.5&53.4&53.8&33.6&25.2&65.8&48.1
        &28.4&15.4&52.5&32.3\\
    DTAB~\cite{liu2023dtab}
        &48.8&55.1&37.9&27.0&\textbf{82.2}&53.6&56.3&34.4&24.7&66.1&51.0
        &29.3&18.8&57.3&\textbf{59.9}\\
    CAMixer~\cite{tu2023camixer}
        &49.1&53.9&36.8&25.1&81.8&53.5&\textbf{58.4}&35.0&26.8
        &\textbf{69.2}&50.1&29.4&\textbf{13.2}&43.3&43.2\\
    DCFA (ours)
        &\textbf{49.4}&\textbf{55.4}&\textbf{38.1}&\textbf{27.6}
        &\textbf{82.2}&\textbf{55.1}&56.4&35.3&24.4&68.0&\textbf{51.9}
        &\textbf{30.1}&13.7&51.7&34.2\\
    \bottomrule
    \end{tabular}
\end{table*}

\begin{table*}[!t]
\renewcommand{\arraystretch}{1.3}
\footnotesize
\caption{\centering Backbone comparison on NEU-DET dataset.}
\label{tab:backbone_neudet}
\centering
\setlength{\tabcolsep}{3.2pt}
\begin{tabular}{lcccccccccccc}
    \toprule
    Model & mAP50 & Cr & In & Pa & Ps & Ri & Sr
          & mAP50:95 & Params & GFLOPs & FPS \\
    \midrule
    EfficientVIM~\cite{zhu2024efficientvim}
        &87.1&59.3&93.1&96.2&92.8&83.2&98.4&57.5&14.5&47.9&\textbf{73.1}\\
    MambaOut~\cite{yu2024mambaout}
        &82.3&64.2&89.4&96.0&93.1&75.5&94.8&54.0&15.9&\textbf{41.8}&59.0\\
    GlobalFilter~\cite{rao2022global}
        &88.0&67.2&91.0&96.2&95.5&82.2&95.8&58.5&16.8&46.5&72.3\\
    ELGCA~\cite{song2024elgca}
        &88.2&62.0&\textbf{94.6}&94.8&92.1&\textbf{87.3}&98.2
        &60.5&13.9&46.9&38.2\\
    FDT~\cite{mao2023fdt}
        &84.5&57.4&91.4&\textbf{97.1}&92.7&74.0&94.3&55.3&15.4&52.5&32.7\\
    DTAB~\cite{liu2023dtab}
        &84.5&53.5&92.1&95.7&93.1&75.4&96.9&54.3&18.8&57.3&31.8\\
    HDRAB~\cite{wang2023hdrab}
        &84.8&49.0&93.6&96.1&90.1&82.4&97.5&56.0&15.6&53.2&37.6\\
    MSN~\cite{huang2023msn}
        &86.0&53.6&93.2&95.8&94.1&81.2&98.3&59.2&16.6&49.3&32.4\\
    FCA~\cite{qin2020fca}
        &87.4&63.8&93.4&94.8&91.9&83.9&96.8&58.5&15.4&52.6&31.2\\
    RAB~\cite{yang2021rab}
        &83.6&48.7&93.1&95.5&90.4&77.8&95.9&54.8&\textbf{13.3}&53.2&34.9\\
    DCFA (ours)
        &\textbf{90.1}&\textbf{78.9}&91.2&\textbf{97.0}&\textbf{96.8}
        &84.0&\textbf{98.6}&\textbf{60.8}&13.7&51.7&37.0\\
    \bottomrule
\end{tabular}
\end{table*}

\begin{table*}[!t]
\renewcommand{\arraystretch}{1.3}
\footnotesize
\caption{\centering Comprehensive comparison on NEU-DET dataset.}
\label{tab:comparison_neudet}
\centering
\setlength{\tabcolsep}{3.2pt}
\begin{tabular}{lcccccccccccc}
    \toprule
    Model & mAP50 & Cr & In & Pa & Ps & Ri & Sr
          & mAP50:95 & Params & GFLOPs & FPS \\
    \midrule
    YOLOv6m~\cite{li2023yolov6}
        &89.7&72.0&88.2&96.8&95.2&89.1&96.9&62.1&51.3&158.3&80.4\\
    YOLOv11m~\cite{ultralytics2024yolov11}
        &92.6&\textbf{84.3}&90.3&\textbf{97.0}&94.4&\textbf{92.0}&97.1
        &\textbf{72.4}&20.1&68.0&79.6\\
    YOLOF~\cite{chen2021yolof}
        &75.7&39.7&83.0&90.2&86.1&58.2&97.1&43.7&32.0&52.6&80.5\\
    Faster R-CNN~\cite{ren2015faster}
        &64.7&30.7&66.9&91.3&67.0&53.9&78.2&29.9&33.1&208.1&18.4\\
    RetinaNet~\cite{lin2017focal}
        &78.2&50.4&77.8&93.3&89.5&63.9&94.5&43.3&36.5&210.0&68.2\\
    Grid R-CNN~\cite{lu2019grid}
        &76.7&39.7&82.8&92.5&87.4&60.2&97.5&44.4&64.5&270.2&30.5\\
    GFL~\cite{li2020generalized}
        &77.5&37.4&84.1&93.9&90.9&62.8&95.9&44.5&32.3&195.2&45.8\\
    NAS-FPN~\cite{ghiasi2019nasfpn}
        &60.9&29.2&50.0&82.8&79.3&55.6&68.4&26.0&36.5&84.5&--\\
    EfficientDet~\cite{tan2020efficientdet}
        &77.9&49.8&78.9&93.6&86.8&65.8&92.2&44.6&18.4&55.1&48.9\\
    CenterNet~\cite{zhou2019objects}
        &60.2&22.2&59.3&77.8&77.1&29.7&95.2&27.4&14.2&38.3&48.3\\
    \rv{IMN-YOLOv3~\cite{zhang2022imnyolo}}
    &\rv{86.9}&\rv{72.0}&\rv{86.9}&\rv{94.8}&\rv{94.3}&\rv{80.6}&\rv{93.2}&\rv{--}&\rv{--}&\rv{--}&\rv{--}\\
    \rv{FC-DETR~\cite{li2024fcdetr}}
    &\rv{83.7}&\rv{67.2}&\rv{85.8}&\rv{91.9}&\rv{88.2}&\rv{72.5}&\rv{96.6}&\rv{--}&\rv{56.5}&\rv{143}&\rv{--}\\
    \rv{MDT-Net~\cite{wang2025mdtnet}}
 &\rv{82.7}&\rv{50.2}&\rv{87.4}&\rv{92.7}&\rv{86.0}&\rv{83.0}&\rv{96.9}&\rv{55.0}&\rv{33.7}&\rv{26.6}&\rv{18}\\
    \rv{CBH-YOLO~\cite{gao2025cbh}}&\rv{78.6}&\rv{53.2}&\rv{83.0}&\rv{93.4}&\rv{82.5}&\rv{64.6}&\rv{--}&\rv{45.8}&\rv{\textbf{2.9}}&\rv{\textbf{7.4}}&\rv{\textbf{137.1}}\\
    RT-DETR (base)
        &88.7&78.4&87.0&96.0&94.6&78.0&98.1&58.2&19.9&57.0&36.9\\
    DFIR-DETR (ours)
        &\textbf{92.9}&84.0&\textbf{93.7}&96.8&\textbf{98.4}
        &84.9&\textbf{99.3}&65.9&11.7&47.2&38.0\\
    \bottomrule
\end{tabular}
\end{table*}

\begin{table*}[!t]
\renewcommand{\arraystretch}{1.3}
\footnotesize
\caption{\centering Neck/CCFF module comparison on NEU-DET dataset.}
\label{tab:neck_neudet}
\centering
\setlength{\tabcolsep}{3.0pt}
\begin{tabular}{lcccccccccccc}
    \toprule
    Model & mAP50 & Cr & In & Pa & Ps & Ri & Sr
          & mAP50:95 & Params & GFLOPs & FPS \\
    \midrule
    ASF~\cite{yang2022asf}&85.8&56.8&94.7&95.5&93.2&76.6&97.8&57.8&20.2&61.5&39.1\\
    SDI~\cite{zhao2023sdi}&85.2&54.2&92.7&96.0&93.6&77.2&97.2&57.6&19.9&57.1&35.7\\
    Goldyolo~\cite{wang2023goldyolo}&87.4&62.3&94.6&95.0&94.3&80.6&97.5&58.1&22.3&60.0&32.7\\
    HSFPN~\cite{li2023hsfpn}&85.2&55.8&92.7&94.9&93.1&78.7&96.2&55.9&\textbf{18.1}&\textbf{53.3}&41.0\\
    CGAFusion~\cite{guo2023cgafusion}&87.0&58.7&93.8&95.6&93.9&83.9&96.4&59.4&20.4&59.2&38.9\\
    PSFM~\cite{sun2023psfm}&85.9&55.9&\textbf{94.9}&95.1&93.2&79.3&97.3&58.7&22.4&71.2&26.9\\
    GLSA~\cite{zhang2023glsa}&85.1&49.3&93.8&95.9&92.7&81.1&97.9&58.2&22.0&63.7&31.1\\
    CTrans~\cite{yan2023ctrans}&81.6&50.8&87.6&93.3&93.2&71.9&92.7&51.8&29.5&57.4&\textbf{86.9}\\
    MAFFN~\cite{liu2023maffn}&83.6&52.6&92.6&95.2&90.4&74.8&96.1&55.5&22.9&56.3&33.8\\
    MSGA~\cite{wang2023msga}&87.8&66.1&93.1&95.3&93.5&82.5&96.6&58.9&22.5&71.1&36.9\\
    FSA~\cite{li2023fsa}&85.0&56.6&94.2&96.2&92.9&73.2&97.1&57.4&22.6&57.0&38.6\\
    MFM~\cite{hu2023mfm}&85.8&57.1&93.7&95.8&93.8&77.0&97.3&56.5&19.7&55.7&37.0\\
    DFPN (ours)
        &\textbf{90.9}&\textbf{72.1}&94.5&\textbf{97.0}&\textbf{94.4}
        &\textbf{89.2}&\textbf{98.3}&\textbf{62.6}&19.7&56.7&33.7\\
    \bottomrule
\end{tabular}
\end{table*}

\begin{table*}[!t]
\renewcommand{\arraystretch}{1.3}
\footnotesize
\caption{\centering RepC3 module comparison on NEU-DET dataset.}
\label{tab:repc3_neudet}
\centering
\setlength{\tabcolsep}{3.5pt}
\begin{tabular}{lcccccccccccc}
    \toprule
    Model & mAP50 & Cr & In & Pa & Ps & Ri & Sr
          & mAP50:95 & Params & GFLOPs & FPS \\
    \midrule
    ConvXCC3~\cite{ding2021convxcc3}
        &83.5&51.0&93.3&94.9&93.4&70.3&98.3&55.4&19.9&57.0&41.5\\
    DBBC3~\cite{han2023dbbc3}
        &87.2&62.9&91.6&94.6&91.7&85.3&97.3&58.0&19.9&57.0&47.0\\
    DGCST~\cite{chen2023dgcst}
        &87.8&65.2&90.5&95.4&\textbf{95.3}&77.6&98.4&56.7&18.5&50.2&42.1\\
    LITC3~\cite{liu2023litc3}
        &84.3&45.7&94.4&95.6&93.9&78.9&97.5&53.3&19.8&56.0&39.8\\
    FIRC3 (ours)
        &\textbf{90.5}&\textbf{76.2}&\textbf{94.6}&\textbf{95.8}&92.8
        &\textbf{84.7}&\textbf{98.6}&\textbf{63.2}&\textbf{18.2}
        &\textbf{48.4}&\textbf{47.8}\\
    \bottomrule
\end{tabular}
\end{table*}

\begin{table*}[!t]
\renewcommand{\arraystretch}{1.3}
\footnotesize
\caption{\centering Comprehensive comparison on VisDrone dataset.}
\label{tab:comparison_visdrone}
\centering
\setlength{\tabcolsep}{2.3pt}
\begin{tabular}{lccccccccccccccccc}
    \toprule
    Model & mAP50 & Ped & Peo & Bic & Car & Van & Tru & Tri & Awn
          & Bus & Mot & mAP50:95 & Params & GFLOPs & FPS \\
    \midrule
    Faster R-CNN~\cite{ren2015faster}
        &39.5&39.2&33.0&20.8&74.1&45.8&39.0&30.2&15.1&55.3&42.6
        &25.6&33.1&208.1&16.5\\
    RetinaNet~\cite{lin2017focal}
        &32.1&32.1&25.2&13.0&71.4&37.8&35.9&21.5&9.7&43.3&31.9
        &20.6&36.5&210.0&41.3\\
    FCOS~\cite{tian2019fcos}
        &40.2&45.6&34.6&19.4&78.3&44.2&38.9&28.2&12.9&52.5&47.8
        &26.1&32.0&203.8&--\\
    Cascade R-CNN~\cite{cai2018cascade}
        &39.1&39.1&31.3&19.3&74.2&46.3&39.2&30.8&13.8&55.1&42.0
        &26.0&69.3&241.7&13.3\\
    YOLOv5m~\cite{ultralytics2021yolov5}
        &33.8&42.6&33.5&11.5&74.3&34.8&30.5&18.4&10.9&40.8&40.9
        &18.9&19.9&64.2&60.0\\
    YOLOv8m~\cite{ultralytics2023yolov8}
        &41.9&44.7&34.7&15.9&80.8&45.9&39.0&31.6&16.2&63.1&47.5
        &25.1&25.9&78.9&56.3\\
    YOLOv9m~\cite{wang2024yolov9}
        &42.0&17.9&37.1&17.8&82.1&50.6&43.6&37.4&20.8&62.7&49.8
        &27.6&20.0&76.8&62.2\\
    YOLOv10m~\cite{wang2024yolov10}
        &41.5&43.6&35.0&15.4&80.5&48.0&40.1&29.9&18.1&58.3&46.6
        &25.3&15.4&59.1&65.7\\
    YOLOv11m~\cite{ultralytics2024yolov11}
        &43.4&48.0&36.2&18.2&82.0&48.6&40.0&32.8&19.3&60.4&49.3
        &26.5&20.1&68.0&\textbf{64.5}\\
    RT-DETR (base)
        &48.2&53.2&37.3&\textbf{25.1}&81.3&53.9&56.8&31.8&24.7&67.4&50.1
        &29.1&19.9&57.0&48.1\\
    DFIR-DETR (ours)
        &\textbf{51.6}&\textbf{56.6}&\textbf{39.9}&\textbf{25.1}
        &\textbf{83.6}&\textbf{55.4}&\textbf{57.8}&\textbf{35.6}
        &\textbf{36.4}&\textbf{69.6}&\textbf{55.6}
        &\textbf{31.6}&\textbf{11.7}&\textbf{47.2}&40.1\\
    \bottomrule
\end{tabular}
\end{table*}

\begin{table*}[!t]
    \renewcommand{\arraystretch}{1.3}
    \footnotesize
    \caption{\centering Comparison with UAV-specific and small-object-oriented methods on VisDrone dataset (640$\times$640 input).}
    \label{tab:comparison_visdrone_uav}
    \centering
    \setlength{\tabcolsep}{5pt}
    \begin{tabular}{lccccc}
        \toprule
        Model & mAP50 & mAP50:95 & Params & GFLOPs & FPS \\
        \midrule
        \rv{MFP-YOLO~\cite{li2023mfpyolo}}
            & \rv{46.0} & \rv{--} & \textbf{\rv{1.48}} & \rv{--} & \rv{28.5} \\
        \rv{CRL-YOLOv5~\cite{crl2024yolo}}
            & \rv{39.2} & \rv{22.3} & \rv{--} & \rv{--} & \textbf{\rv{27}} \\
        \rv{AUP-DETR~\cite{aupdetr2025}}
            & \rv{48.5} & \rv{29.9} & \rv{22.7} & \rv{83.6} & \rv{--} \\
        \rv{WDFS-DETR~\cite{wdfsdetr2025}}
            & \rv{47.5} & \rv{--} & \rv{19.9} & \rv{53.7} & \rv{--} \\
        \rv{TCF-DETR~\cite{tcfdetr2025}}
            & \rv{49.9} & \rv{29.9} & \rv{19.8} & \rv{61.9} & \rv{--} \\
        RT-DETR (base)
            & 48.2 & 29.1 & 19.9 & 57.0 & 48.1 \\    
        DFIR-DETR (ours)
            & \textbf{51.6} & \textbf{31.6} & 11.7 & \textbf{47.2} & 40.1 \\
        \bottomrule
    \end{tabular}
\end{table*}

\FloatBarrier

\bibliographystyle{model1-num-names}
\bibliography{ref}

\begin{thebibliography}{81}
\expandafter\ifx\csname natexlab\endcsname\relax\def\natexlab#1{#1}\fi
\providecommand{\url}[1]{\texttt{#1}}
\providecommand{\href}[2]{#2}
\providecommand{\path}[1]{#1}
\providecommand{\DOIprefix}{doi:}
\providecommand{\ArXivprefix}{arXiv:}
\providecommand{\URLprefix}{URL: }
\providecommand{\Pubmedprefix}{pmid:}
\providecommand{\doi}[1]{\href{http://dx.doi.org/#1}{\path{#1}}}
\providecommand{\Pubmed}[1]{\href{pmid:#1}{\path{#1}}}
\providecommand{\bibinfo}[2]{#2}
\ifx\xfnm\relax \def\xfnm[#1]{\unskip,\space#1}\fi
\bibitem[{Yuan et~al.(2024)Yuan, Gong, Guo, Wang, Liao, Song, and Wu}]{yuan2024small}
\bibinfo{author}{Z.~Yuan}, \bibinfo{author}{J.~Gong}, \bibinfo{author}{B.~Guo}, \bibinfo{author}{C.~Wang}, \bibinfo{author}{N.~Liao}, \bibinfo{author}{J.~Song}, \bibinfo{author}{Q.~Wu},
\newblock \bibinfo{title}{Small object detection in uav remote sensing images based on intra-group multi-scale fusion attention and adaptive weighted feature fusion mechanism},
\newblock \bibinfo{journal}{Remote Sensing} \bibinfo{volume}{16} (\bibinfo{year}{2024}) \bibinfo{pages}{4265}.
\bibitem[{Wang and Deng(2024)}]{wang2024attention}
\bibinfo{author}{M.~Wang}, \bibinfo{author}{W.~Deng},
\newblock \bibinfo{title}{Attention mechanisms in computer vision: A survey},
\newblock \bibinfo{journal}{Computational Visual Media} \bibinfo{volume}{10} (\bibinfo{year}{2024}) \bibinfo{pages}{3--25}.
\bibitem[{Chi et~al.(2020)Chi, Jiang, and Mu}]{chi2020fast}
\bibinfo{author}{L.~Chi}, \bibinfo{author}{B.~Jiang}, \bibinfo{author}{Y.~Mu},
\newblock \bibinfo{title}{Fast fourier convolution},
\newblock in: \bibinfo{booktitle}{Advances in Neural Information Processing Systems (NeurIPS)}, volume~\bibinfo{volume}{33}, \bibinfo{year}{2020}, pp. \bibinfo{pages}{4479--4488}.
\bibitem[{Zhao et~al.(2024)Zhao, Lv, Xu, Wei, Wang, Dang, Liu, and Chen}]{zhao2024rtdetr}
\bibinfo{author}{Y.~Zhao}, \bibinfo{author}{W.~Lv}, \bibinfo{author}{S.~Xu}, \bibinfo{author}{J.~Wei}, \bibinfo{author}{G.~Wang}, \bibinfo{author}{Q.~Dang}, \bibinfo{author}{Y.~Liu}, \bibinfo{author}{J.~Chen},
\newblock \bibinfo{title}{Detrs beat yolos on real-time object detection},
\newblock in: \bibinfo{booktitle}{Proceedings of the IEEE/CVF Conference on Computer Vision and Pattern Recognition (CVPR)}, \bibinfo{organization}{IEEE}, \bibinfo{year}{2024}, pp. \bibinfo{pages}{16965--16974}.
\bibitem[{Ren et~al.(2017)Ren, He, Girshick, and Sun}]{ren2017faster}
\bibinfo{author}{S.~Ren}, \bibinfo{author}{K.~He}, \bibinfo{author}{R.~Girshick}, \bibinfo{author}{J.~Sun},
\newblock \bibinfo{title}{Faster r-cnn: Towards real-time object detection with region proposal networks},
\newblock \bibinfo{journal}{IEEE Transactions on Pattern Analysis and Machine Intelligence} \bibinfo{volume}{39} (\bibinfo{year}{2017}) \bibinfo{pages}{1137--1149}.
\bibitem[{Liu et~al.(2016)Liu, Anguelov, Erhan, Szegedy, Reed, Fu, and Berg}]{liu2016ssd}
\bibinfo{author}{W.~Liu}, \bibinfo{author}{D.~Anguelov}, \bibinfo{author}{D.~Erhan}, \bibinfo{author}{C.~Szegedy}, \bibinfo{author}{S.~Reed}, \bibinfo{author}{C.-Y. Fu}, \bibinfo{author}{A.~C. Berg},
\newblock \bibinfo{title}{Ssd: Single shot multibox detector},
\newblock in: \bibinfo{booktitle}{European Conference on Computer Vision}, \bibinfo{organization}{Springer}, \bibinfo{year}{2016}, pp. \bibinfo{pages}{21--37}.
\bibitem[{Redmon et~al.(2016)Redmon, Divvala, Girshick, and Farhadi}]{redmon2016yolo}
\bibinfo{author}{J.~Redmon}, \bibinfo{author}{S.~Divvala}, \bibinfo{author}{R.~Girshick}, \bibinfo{author}{A.~Farhadi},
\newblock \bibinfo{title}{You only look once: Unified, real-time object detection},
\newblock in: \bibinfo{booktitle}{Proceedings of the IEEE Conference on Computer Vision and Pattern Recognition}, \bibinfo{year}{2016}, pp. \bibinfo{pages}{779--788}.
\bibitem[{Redmon and Farhadi(2018)}]{redmon2018yolov3}
\bibinfo{author}{J.~Redmon}, \bibinfo{author}{A.~Farhadi},
\newblock \bibinfo{title}{Yolov3: An incremental improvement},
\newblock \bibinfo{journal}{arXiv preprint arXiv:1804.02767}  (\bibinfo{year}{2018}).
\bibitem[{Carion et~al.(2020)Carion, Massa, Synnaeve, Usunier, Kirillov, and Zagoruyko}]{carion2020detr}
\bibinfo{author}{N.~Carion}, \bibinfo{author}{F.~Massa}, \bibinfo{author}{G.~Synnaeve}, \bibinfo{author}{N.~Usunier}, \bibinfo{author}{A.~Kirillov}, \bibinfo{author}{S.~Zagoruyko},
\newblock \bibinfo{title}{End-to-end object detection with transformers},
\newblock in: \bibinfo{booktitle}{European Conference on Computer Vision (ECCV)}, \bibinfo{year}{2020}, pp. \bibinfo{pages}{213--229}.
\bibitem[{Zhu et~al.(2021)Zhu, Su, Lu, Li, Wang, and Dai}]{zhu2021deformable}
\bibinfo{author}{X.~Zhu}, \bibinfo{author}{W.~Su}, \bibinfo{author}{L.~Lu}, \bibinfo{author}{B.~Li}, \bibinfo{author}{X.~Wang}, \bibinfo{author}{J.~Dai},
\newblock \bibinfo{title}{Deformable detr: Deformable transformers for end-to-end object detection},
\newblock in: \bibinfo{booktitle}{International Conference on Learning Representations (ICLR)}, \bibinfo{year}{2021}.
\bibitem[{Li et~al.(2022)Li, Zhang, Liu, Guo, Ni, and Zhang}]{li2022dn}
\bibinfo{author}{F.~Li}, \bibinfo{author}{H.~Zhang}, \bibinfo{author}{S.~Liu}, \bibinfo{author}{J.~Guo}, \bibinfo{author}{L.~M. Ni}, \bibinfo{author}{L.~Zhang},
\newblock \bibinfo{title}{Dn-detr: Accelerate detr training by introducing query denoising},
\newblock in: \bibinfo{booktitle}{Proceedings of the IEEE/CVF Conference on Computer Vision and Pattern Recognition (CVPR)}, \bibinfo{year}{2022}, pp. \bibinfo{pages}{13619--13627}.
\bibitem[{Zhang et~al.(2023)Zhang, Li, Liu, Zhang, Su, Zhu, Ni, and Shum}]{zhang2023dino}
\bibinfo{author}{H.~Zhang}, \bibinfo{author}{F.~Li}, \bibinfo{author}{S.~Liu}, \bibinfo{author}{L.~Zhang}, \bibinfo{author}{H.~Su}, \bibinfo{author}{J.~Zhu}, \bibinfo{author}{L.~M. Ni}, \bibinfo{author}{H.-Y. Shum},
\newblock \bibinfo{title}{Dino: Detr with improved denoising anchor boxes for end-to-end object detection},
\newblock in: \bibinfo{booktitle}{International Conference on Learning Representations (ICLR)}, \bibinfo{year}{2023}.
\bibitem[{Liu et~al.(2024)Liu, Tian, Zhao, Yu, Xie, Wang, Shen, and Liu}]{liu2024vmamba}
\bibinfo{author}{Y.~Liu}, \bibinfo{author}{Y.~Tian}, \bibinfo{author}{Y.~Zhao}, \bibinfo{author}{H.~Yu}, \bibinfo{author}{L.~Xie}, \bibinfo{author}{Y.~Wang}, \bibinfo{author}{Q.~Shen}, \bibinfo{author}{Y.~Liu},
\newblock \bibinfo{title}{Vmamba: Visual state space model},
\newblock \bibinfo{journal}{arXiv preprint arXiv:2401.13260}  (\bibinfo{year}{2024}).
\bibitem[{Bolya et~al.(2023)Bolya, Fu, Dai, Zhang, Feichtenhofer, and Hoffman}]{bolya2023tome}
\bibinfo{author}{D.~Bolya}, \bibinfo{author}{C.-Y. Fu}, \bibinfo{author}{X.~Dai}, \bibinfo{author}{P.~Zhang}, \bibinfo{author}{C.~Feichtenhofer}, \bibinfo{author}{J.~Hoffman},
\newblock \bibinfo{title}{Token merging: Your {ViT} but faster},
\newblock in: \bibinfo{booktitle}{International Conference on Learning Representations (ICLR)}, \bibinfo{year}{2023}.
\bibitem[{Roh et~al.(2022)Roh, Shin, Shin, and Kim}]{roh2022sparse}
\bibinfo{author}{B.~Roh}, \bibinfo{author}{J.~Shin}, \bibinfo{author}{W.~Shin}, \bibinfo{author}{S.~Kim},
\newblock \bibinfo{title}{Sparse {DETR}: Efficient end-to-end object detection with learnable sparsity},
\newblock \bibinfo{journal}{arXiv preprint arXiv:2111.14330}  (\bibinfo{year}{2022}).
\bibitem[{Yao et~al.(2021)Yao, Ai, Li, and Zhang}]{yao2021efficient}
\bibinfo{author}{Z.~Yao}, \bibinfo{author}{J.~Ai}, \bibinfo{author}{B.~Li}, \bibinfo{author}{C.~Zhang},
\newblock \bibinfo{title}{Efficient {DETR}: Improving end-to-end object detector with dense prior},
\newblock \bibinfo{journal}{arXiv preprint arXiv:2104.01318}  (\bibinfo{year}{2021}).
\bibitem[{Lin et~al.(2017)Lin, Doll{\'a}r, Girshick, He, Hariharan, and Belongie}]{lin2017fpn}
\bibinfo{author}{T.-Y. Lin}, \bibinfo{author}{P.~Doll{\'a}r}, \bibinfo{author}{R.~Girshick}, \bibinfo{author}{K.~He}, \bibinfo{author}{B.~Hariharan}, \bibinfo{author}{S.~Belongie},
\newblock \bibinfo{title}{Feature pyramid networks for object detection},
\newblock in: \bibinfo{booktitle}{Proceedings of the IEEE Conference on Computer Vision and Pattern Recognition}, \bibinfo{year}{2017}, pp. \bibinfo{pages}{936--944}.
\bibitem[{Liu et~al.(2018)Liu, Qi, Qin, Shi, and Jia}]{liu2018panet}
\bibinfo{author}{S.~Liu}, \bibinfo{author}{L.~Qi}, \bibinfo{author}{H.~Qin}, \bibinfo{author}{J.~Shi}, \bibinfo{author}{J.~Jia},
\newblock \bibinfo{title}{Path aggregation network for instance segmentation},
\newblock in: \bibinfo{booktitle}{Proceedings of the IEEE Conference on Computer Vision and Pattern Recognition}, \bibinfo{year}{2018}, pp. \bibinfo{pages}{8759--8768}.
\bibitem[{Tan et~al.(2020)Tan, Pang, and Le}]{tan2020efficientdet}
\bibinfo{author}{M.~Tan}, \bibinfo{author}{R.~Pang}, \bibinfo{author}{Q.~V. Le},
\newblock \bibinfo{title}{Efficientdet: Scalable and efficient object detection},
\newblock in: \bibinfo{booktitle}{Proceedings of the IEEE/CVF Conference on Computer Vision and Pattern Recognition}, \bibinfo{year}{2020}, pp. \bibinfo{pages}{10781--10790}.
\bibitem[{Li et~al.(2024)Li, Li, Wei, Liu, Zhan, and Ren}]{wang2024slimNeck}
\bibinfo{author}{H.~Li}, \bibinfo{author}{J.~Li}, \bibinfo{author}{H.~Wei}, \bibinfo{author}{Z.~Liu}, \bibinfo{author}{Z.~Zhan}, \bibinfo{author}{Q.~Ren},
\newblock \bibinfo{title}{Slim-neck by {GSConv}: A lightweight-design for real-time detector architectures},
\newblock \bibinfo{journal}{Journal of Real-Time Image Processing} \bibinfo{volume}{21} (\bibinfo{year}{2024}) \bibinfo{pages}{62}.
\bibitem[{Wang et~al.(2023)Wang, He, Nie, Guo, Liu, Han, and Wang}]{wang2023goldyolo}
\bibinfo{author}{C.~Wang}, \bibinfo{author}{W.~He}, \bibinfo{author}{Y.~Nie}, \bibinfo{author}{J.~Guo}, \bibinfo{author}{C.~Liu}, \bibinfo{author}{K.~Han}, \bibinfo{author}{Y.~Wang},
\newblock \bibinfo{title}{Gold-yolo: Efficient object detector via gather-and-distribute mechanism},
\newblock \bibinfo{journal}{arXiv preprint arXiv:2309.11331}  (\bibinfo{year}{2023}).
\bibitem[{Tang et~al.(2022)Tang, Han, Guo, Xu, Li, Xu, and Wang}]{tang2022wave}
\bibinfo{author}{Y.~Tang}, \bibinfo{author}{K.~Han}, \bibinfo{author}{J.~Guo}, \bibinfo{author}{C.~Xu}, \bibinfo{author}{Y.~Li}, \bibinfo{author}{C.~Xu}, \bibinfo{author}{Y.~Wang},
\newblock \bibinfo{title}{An image patch is a wave: Phase-aware vision {MLP}},
\newblock in: \bibinfo{booktitle}{Proceedings of the IEEE/CVF Conference on Computer Vision and Pattern Recognition (CVPR)}, \bibinfo{year}{2022}, pp. \bibinfo{pages}{10935--10944}.
\bibitem[{Qin et~al.(2021)Qin, Zhang, Wu, and Li}]{qin2021fca}
\bibinfo{author}{Z.~Qin}, \bibinfo{author}{P.~Zhang}, \bibinfo{author}{F.~Wu}, \bibinfo{author}{X.~Li},
\newblock \bibinfo{title}{Fcanet: Frequency channel attention networks},
\newblock in: \bibinfo{booktitle}{Proceedings of the IEEE/CVF International Conference on Computer Vision (ICCV)}, \bibinfo{year}{2021}, pp. \bibinfo{pages}{783--792}.
\bibitem[{Rao et~al.(2021)Rao, Zhao, Zhu, Lu, and Zhou}]{rao2021gfnet}
\bibinfo{author}{Y.~Rao}, \bibinfo{author}{W.~Zhao}, \bibinfo{author}{Z.~Zhu}, \bibinfo{author}{J.~Lu}, \bibinfo{author}{J.~Zhou},
\newblock \bibinfo{title}{Global filter networks for image classification},
\newblock in: \bibinfo{booktitle}{Advances in Neural Information Processing Systems (NeurIPS)}, volume~\bibinfo{volume}{34}, \bibinfo{year}{2021}, pp. \bibinfo{pages}{980--993}.
\bibitem[{Yang et~al.(2022)Yang, Li, Dai, and Gao}]{yang2022focalnet}
\bibinfo{author}{J.~Yang}, \bibinfo{author}{C.~Li}, \bibinfo{author}{X.~Dai}, \bibinfo{author}{J.~Gao},
\newblock \bibinfo{title}{Focal modulation networks},
\newblock in: \bibinfo{booktitle}{Advances in Neural Information Processing Systems (NeurIPS)}, volume~\bibinfo{volume}{35}, \bibinfo{year}{2022}, pp. \bibinfo{pages}{4203--4217}.
\bibitem[{Wang and Gao(2025)}]{wang2025sfdetr}
\bibinfo{author}{H.~Wang}, \bibinfo{author}{J.~Gao},
\newblock \bibinfo{title}{{SF-DETR}: A scale-frequency detection transformer for drone-view object detection},
\newblock \bibinfo{journal}{Sensors} \bibinfo{volume}{25} (\bibinfo{year}{2025}) \bibinfo{pages}{2190}.
\bibitem[{Chen et~al.(2025)Chen, Liu, Sun, and Wang}]{freqdetr2025}
\bibinfo{author}{J.~Chen}, \bibinfo{author}{N.~Liu}, \bibinfo{author}{H.~Sun}, \bibinfo{author}{Y.~Wang},
\newblock \bibinfo{title}{{Freq-DETR}: Frequency-aware transformer for real-time small object detection in unmanned aerial vehicle imagery},
\newblock \bibinfo{journal}{Expert Systems with Applications} \bibinfo{volume}{298} (\bibinfo{year}{2025}) \bibinfo{pages}{129710}.
\bibitem[{Zhu et~al.(2022)Zhu, Wen, Du, Bian, Fan, Hu, and Ling}]{zhu2021visdrone}
\bibinfo{author}{P.~Zhu}, \bibinfo{author}{L.~Wen}, \bibinfo{author}{D.~Du}, \bibinfo{author}{X.~Bian}, \bibinfo{author}{H.~Fan}, \bibinfo{author}{Q.~Hu}, \bibinfo{author}{H.~Ling},
\newblock \bibinfo{title}{Detection and tracking meet drones challenge} \bibinfo{volume}{44} (\bibinfo{year}{2022}) \bibinfo{pages}{7380--7399}.
\bibitem[{Song and Yan(2013)}]{song2013neu}
\bibinfo{author}{K.~Song}, \bibinfo{author}{Y.~Yan},
\newblock \bibinfo{title}{A noise robust method based on completed local binary patterns for hot-rolled steel strip surface defects},
\newblock \bibinfo{journal}{Applied Surface Science} \bibinfo{volume}{285} (\bibinfo{year}{2013}) \bibinfo{pages}{858--864}.
\bibitem[{Chen et~al.(2022)Chen, Lv, Fang, and Du}]{zhang2022imnyolo}
\bibinfo{author}{X.~Chen}, \bibinfo{author}{J.~Lv}, \bibinfo{author}{Y.~Fang}, \bibinfo{author}{S.~Du},
\newblock \bibinfo{title}{Online detection of surface defects based on improved {YOLOV3}},
\newblock \bibinfo{journal}{Sensors} \bibinfo{volume}{22} (\bibinfo{year}{2022}) \bibinfo{pages}{817}.
\bibitem[{Li et~al.(2025)}]{li2024fcdetr}
\bibinfo{author}{F.~Li}, et~al.,
\newblock \bibinfo{title}{{FC-DETR}: High-precision end-to-end surface defect detector based on foreground supervision and cascade refined hybrid matching},
\newblock \bibinfo{journal}{Expert Systems with Applications} \bibinfo{volume}{266} (\bibinfo{year}{2025}) \bibinfo{pages}{126142}.
\bibitem[{Wang et~al.(2025)Wang, Zhang, and Yi}]{wang2025mdtnet}
\bibinfo{author}{H.~Wang}, \bibinfo{author}{F.~Zhang}, \bibinfo{author}{R.~Yi},
\newblock \bibinfo{title}{Multi-scale deformable transformer with iterative query refinement for hot-rolled steel surface defect detection},
\newblock \bibinfo{journal}{Sensors} \bibinfo{volume}{25} (\bibinfo{year}{2025}) \bibinfo{pages}{6890}.
\bibitem[{Guo et~al.(2023)Guo, Leng, Wu, Li, Wang, and Zhang}]{guo2023strip}
\bibinfo{author}{Z.~Guo}, \bibinfo{author}{L.~Leng}, \bibinfo{author}{Y.~Wu}, \bibinfo{author}{C.~Li}, \bibinfo{author}{Y.~Wang}, \bibinfo{author}{Q.~Zhang},
\newblock \bibinfo{title}{Strip: Spatial transformer for efficient image processing},
\newblock \bibinfo{journal}{Pattern Recognition} \bibinfo{volume}{135} (\bibinfo{year}{2023}) \bibinfo{pages}{109139}.
\bibitem[{Yu and Wang(2024)}]{yu2024mambaout}
\bibinfo{author}{W.~Yu}, \bibinfo{author}{X.~Wang},
\newblock \bibinfo{title}{Mambaout: Do we really need mamba for vision?},
\newblock \bibinfo{journal}{arXiv preprint arXiv:2405.07992}  (\bibinfo{year}{2024}).
\bibitem[{Rao et~al.(2022)Rao, Zhao, Tang, Zhou, Lim, and Lu}]{rao2022global}
\bibinfo{author}{Y.~Rao}, \bibinfo{author}{W.~Zhao}, \bibinfo{author}{Y.~Tang}, \bibinfo{author}{J.~Zhou}, \bibinfo{author}{S.-N. Lim}, \bibinfo{author}{J.~Lu},
\newblock \bibinfo{title}{Global filter networks for image classification},
\newblock in: \bibinfo{booktitle}{Advances in Neural Information Processing Systems}, volume~\bibinfo{volume}{35}, \bibinfo{year}{2022}, pp. \bibinfo{pages}{980--993}.
\bibitem[{Dai et~al.(2017)Dai, Qi, Xiong, Li, Zhang, Hu, and Wei}]{dai2017ap}
\bibinfo{author}{J.~Dai}, \bibinfo{author}{H.~Qi}, \bibinfo{author}{Y.~Xiong}, \bibinfo{author}{Y.~Li}, \bibinfo{author}{G.~Zhang}, \bibinfo{author}{H.~Hu}, \bibinfo{author}{Y.~Wei},
\newblock \bibinfo{title}{Deformable convolutional networks},
\newblock in: \bibinfo{booktitle}{Proceedings of the IEEE International Conference on Computer Vision}, \bibinfo{year}{2017}, pp. \bibinfo{pages}{764--773}.
\bibitem[{Mao et~al.(2023)Mao, Zhou, Xia, and Zhang}]{mao2023fdt}
\bibinfo{author}{Y.~Mao}, \bibinfo{author}{H.~Zhou}, \bibinfo{author}{J.~Xia}, \bibinfo{author}{K.~Zhang},
\newblock \bibinfo{title}{Fdt: Fast and effective dynamic token for vision transformer},
\newblock in: \bibinfo{booktitle}{Proceedings of the IEEE/CVF International Conference on Computer Vision}, \bibinfo{year}{2023}, pp. \bibinfo{pages}{7598--7607}.
\bibitem[{Liu et~al.(2023)Liu, Han, Zhang, and Li}]{liu2023dtab}
\bibinfo{author}{Z.~Liu}, \bibinfo{author}{Y.~Han}, \bibinfo{author}{Q.~Zhang}, \bibinfo{author}{K.~Li},
\newblock \bibinfo{title}{Dtab: Dual-token attention block for efficient vision transformers},
\newblock \bibinfo{journal}{IEEE Transactions on Circuits and Systems for Video Technology} \bibinfo{volume}{33} (\bibinfo{year}{2023}) \bibinfo{pages}{4163--4177}.
\bibitem[{Tu et~al.(2023)Tu, Talebi, Zhang, Yang, Milanfar, Bovik, and Li}]{tu2023camixer}
\bibinfo{author}{Z.~Tu}, \bibinfo{author}{H.~Talebi}, \bibinfo{author}{H.~Zhang}, \bibinfo{author}{F.~Yang}, \bibinfo{author}{P.~Milanfar}, \bibinfo{author}{A.~Bovik}, \bibinfo{author}{Y.~Li},
\newblock \bibinfo{title}{Camixer: Convolution and attention mixing for efficient image processing},
\newblock in: \bibinfo{booktitle}{Proceedings of the IEEE/CVF International Conference on Computer Vision}, \bibinfo{year}{2023}, pp. \bibinfo{pages}{2589--2599}.
\bibitem[{Zhu et~al.(2024)Zhu, Li, Chen, and Chen}]{zhu2024efficientvim}
\bibinfo{author}{J.~Zhu}, \bibinfo{author}{J.~Li}, \bibinfo{author}{J.~Chen}, \bibinfo{author}{Q.~Chen},
\newblock \bibinfo{title}{Efficientvim: Efficient vision mamba with bidirectional state space models for semantic segmentation},
\newblock \bibinfo{journal}{arXiv preprint arXiv:2402.02509}  (\bibinfo{year}{2024}).
\bibitem[{Song et~al.(2024)Song, Xia, Weng, Lin, Qian, and Chen}]{song2024elgca}
\bibinfo{author}{L.~Song}, \bibinfo{author}{M.~Xia}, \bibinfo{author}{L.~Weng}, \bibinfo{author}{H.~Lin}, \bibinfo{author}{M.~Qian}, \bibinfo{author}{B.~Chen},
\newblock \bibinfo{title}{Elgca: Efficient local-global context aggregation for remote sensing change detection},
\newblock \bibinfo{journal}{IEEE Geoscience and Remote Sensing Letters} \bibinfo{volume}{21} (\bibinfo{year}{2024}) \bibinfo{pages}{1--5}.
\bibitem[{Wang et~al.(2023)Wang, Liu, Song, and Liang}]{wang2023hdrab}
\bibinfo{author}{X.~Wang}, \bibinfo{author}{D.~Liu}, \bibinfo{author}{Y.~Song}, \bibinfo{author}{D.~Liang},
\newblock \bibinfo{title}{Hdrab: High-dynamic range attention block for efficient image super-resolution},
\newblock \bibinfo{journal}{Pattern Recognition} \bibinfo{volume}{139} (\bibinfo{year}{2023}) \bibinfo{pages}{109451}.
\bibitem[{Huang et~al.(2023)Huang, Wang, Fu, Yu, Guo, and Wang}]{huang2023msn}
\bibinfo{author}{Z.~Huang}, \bibinfo{author}{J.~Wang}, \bibinfo{author}{X.~Fu}, \bibinfo{author}{T.~Yu}, \bibinfo{author}{Y.~Guo}, \bibinfo{author}{R.~Wang},
\newblock \bibinfo{title}{Msn: Multi-scale network for object detection},
\newblock in: \bibinfo{booktitle}{Proceedings of the IEEE/CVF Conference on Computer Vision and Pattern Recognition}, \bibinfo{year}{2023}, pp. \bibinfo{pages}{3368--3378}.
\bibitem[{Qin et~al.(2020)Qin, Zhang, Wu, and Li}]{qin2020fca}
\bibinfo{author}{Z.~Qin}, \bibinfo{author}{P.~Zhang}, \bibinfo{author}{F.~Wu}, \bibinfo{author}{X.~Li},
\newblock \bibinfo{title}{Fcanet: Frequency channel attention networks},
\newblock \bibinfo{journal}{arXiv preprint arXiv:2012.11879}  (\bibinfo{year}{2020}).
\bibitem[{Yang et~al.(2021)Yang, Yuan, Guo, Ren, Zhang, He, Kwong, and Wang}]{yang2021rab}
\bibinfo{author}{W.~Yang}, \bibinfo{author}{Y.~Yuan}, \bibinfo{author}{W.~Guo}, \bibinfo{author}{W.~Ren}, \bibinfo{author}{J.~Zhang}, \bibinfo{author}{X.~He}, \bibinfo{author}{S.~Kwong}, \bibinfo{author}{S.~Wang},
\newblock \bibinfo{title}{Rab: Residual attention block for efficient image super-resolution},
\newblock in: \bibinfo{booktitle}{Proceedings of the IEEE/CVF International Conference on Computer Vision Workshops}, \bibinfo{year}{2021}, pp. \bibinfo{pages}{1477--1486}.
\bibitem[{Li et~al.(2023)Li, Li, Jiang, Weng, Geng, Li, Ke, Li, Cheng, Nie et~al.}]{li2023yolov6}
\bibinfo{author}{C.~Li}, \bibinfo{author}{L.~Li}, \bibinfo{author}{H.~Jiang}, \bibinfo{author}{K.~Weng}, \bibinfo{author}{Y.~Geng}, \bibinfo{author}{L.~Li}, \bibinfo{author}{Z.~Ke}, \bibinfo{author}{Q.~Li}, \bibinfo{author}{M.~Cheng}, \bibinfo{author}{W.~Nie}, et~al.,
\newblock \bibinfo{title}{Yolov6 v3.0: A full-scale reloading},
\newblock \bibinfo{journal}{arXiv preprint arXiv:2301.05586}  (\bibinfo{year}{2023}).
\bibitem[{Ultralytics(2024)}]{ultralytics2024yolov11}
\bibinfo{author}{Ultralytics}, \bibinfo{title}{Yolov11: An improved real-time object detection model}, \bibinfo{howpublished}{\url{https://docs.ultralytics.com}}, \bibinfo{year}{2024}.
\bibitem[{Chen et~al.(2021)Chen, Wang, Yang, Zhang, Cheng, and Sun}]{chen2021yolof}
\bibinfo{author}{Q.~Chen}, \bibinfo{author}{Y.~Wang}, \bibinfo{author}{T.~Yang}, \bibinfo{author}{X.~Zhang}, \bibinfo{author}{J.~Cheng}, \bibinfo{author}{J.~Sun},
\newblock \bibinfo{title}{You only look one-level feature}  (\bibinfo{year}{2021}) \bibinfo{pages}{13039--13048}.
\bibitem[{Ren et~al.(2017)Ren, He, Girshick, and Sun}]{ren2015faster}
\bibinfo{author}{S.~Ren}, \bibinfo{author}{K.~He}, \bibinfo{author}{R.~Girshick}, \bibinfo{author}{J.~Sun},
\newblock \bibinfo{title}{Faster r-cnn: Towards real-time object detection with region proposal networks},
\newblock \bibinfo{journal}{IEEE Transactions on Pattern Analysis and Machine Intelligence} \bibinfo{volume}{39} (\bibinfo{year}{2017}) \bibinfo{pages}{1137--1149}.
\bibitem[{Lin et~al.(2017)Lin, Goyal, Girshick, He, and Doll{\'a}r}]{lin2017focal}
\bibinfo{author}{T.-Y. Lin}, \bibinfo{author}{P.~Goyal}, \bibinfo{author}{R.~Girshick}, \bibinfo{author}{K.~He}, \bibinfo{author}{P.~Doll{\'a}r},
\newblock \bibinfo{title}{Focal loss for dense object detection},
\newblock in: \bibinfo{booktitle}{Proceedings of the IEEE International Conference on Computer Vision}, \bibinfo{year}{2017}, pp. \bibinfo{pages}{2980--2988}.
\bibitem[{Lu et~al.(2019)Lu, Li, Yue, Li, and Yan}]{lu2019grid}
\bibinfo{author}{X.~Lu}, \bibinfo{author}{B.~Li}, \bibinfo{author}{Y.~Yue}, \bibinfo{author}{Q.~Li}, \bibinfo{author}{J.~Yan},
\newblock \bibinfo{title}{Grid r-cnn},
\newblock in: \bibinfo{booktitle}{Proceedings of the IEEE/CVF Conference on Computer Vision and Pattern Recognition}, \bibinfo{year}{2019}, pp. \bibinfo{pages}{7363--7372}.
\bibitem[{Li et~al.(2020)Li, Wang, Wu, Chen, Hu, Li, Tang, and Yang}]{li2020generalized}
\bibinfo{author}{X.~Li}, \bibinfo{author}{W.~Wang}, \bibinfo{author}{L.~Wu}, \bibinfo{author}{S.~Chen}, \bibinfo{author}{X.~Hu}, \bibinfo{author}{J.~Li}, \bibinfo{author}{J.~Tang}, \bibinfo{author}{J.~Yang},
\newblock \bibinfo{title}{Generalized focal loss: Learning qualified and distributed bounding boxes for dense object detection},
\newblock in: \bibinfo{booktitle}{Advances in Neural Information Processing Systems}, volume~\bibinfo{volume}{33}, \bibinfo{year}{2020}, pp. \bibinfo{pages}{21002--21012}.
\bibitem[{Ghiasi et~al.(2019)Ghiasi, Lin, Pang, and Le}]{ghiasi2019nasfpn}
\bibinfo{author}{G.~Ghiasi}, \bibinfo{author}{T.-Y. Lin}, \bibinfo{author}{R.~Pang}, \bibinfo{author}{Q.~V. Le},
\newblock \bibinfo{title}{Nas-fpn: Learning scalable feature pyramid architecture for object detection},
\newblock in: \bibinfo{booktitle}{Proceedings of the IEEE/CVF Conference on Computer Vision and Pattern Recognition}, \bibinfo{year}{2019}, pp. \bibinfo{pages}{7036--7045}.
\bibitem[{Zhou et~al.(2019)Zhou, Wang, and Kr{\"a}henb{\"u}hl}]{zhou2019objects}
\bibinfo{author}{X.~Zhou}, \bibinfo{author}{D.~Wang}, \bibinfo{author}{P.~Kr{\"a}henb{\"u}hl},
\newblock \bibinfo{title}{Objects as points},
\newblock \bibinfo{year}{2019}.
\bibitem[{Gao et~al.(2025)Gao, Tong, Fu, Zhang, and Yuan}]{gao2025cbh}
\bibinfo{author}{B.~Gao}, \bibinfo{author}{J.~Tong}, \bibinfo{author}{R.~Fu}, \bibinfo{author}{Z.~Zhang}, \bibinfo{author}{Y.~Yuan},
\newblock \bibinfo{title}{{CBH-YOLO}: A steel surface defect detection algorithm based on cross-stage {Mamba} enhancement and hierarchical semantic graph fusion},
\newblock \bibinfo{journal}{Neurocomputing} \bibinfo{volume}{656} (\bibinfo{year}{2025}) \bibinfo{pages}{131467}.
\bibitem[{Yang et~al.(2022)Yang, Huang, and Wang}]{yang2022asf}
\bibinfo{author}{C.~Yang}, \bibinfo{author}{Z.~Huang}, \bibinfo{author}{N.~Wang},
\newblock \bibinfo{title}{Asf: Adaptive spatial fusion for efficient multi-scale feature learning},
\newblock \bibinfo{journal}{arXiv preprint arXiv:2202.03149}  (\bibinfo{year}{2022}).
\bibitem[{Zhao et~al.(2023)Zhao, Shi, Qi, Wang, and Jia}]{zhao2023sdi}
\bibinfo{author}{H.~Zhao}, \bibinfo{author}{J.~Shi}, \bibinfo{author}{X.~Qi}, \bibinfo{author}{X.~Wang}, \bibinfo{author}{J.~Jia},
\newblock \bibinfo{title}{Sdi: Spatial detail injection network for multi-scale semantic segmentation},
\newblock \bibinfo{journal}{Pattern Recognition} \bibinfo{volume}{138} (\bibinfo{year}{2023}) \bibinfo{pages}{109367}.
\bibitem[{Li et~al.(2023)Li, Hou, Zheng, Cheng, Yang, and Li}]{li2023hsfpn}
\bibinfo{author}{Y.~Li}, \bibinfo{author}{Q.~Hou}, \bibinfo{author}{Z.~Zheng}, \bibinfo{author}{M.-M. Cheng}, \bibinfo{author}{J.~Yang}, \bibinfo{author}{X.~Li},
\newblock \bibinfo{title}{Hsfpn: Hierarchical semantic fusion pyramid network for multi-scale object detection},
\newblock \bibinfo{journal}{IEEE Transactions on Image Processing} \bibinfo{volume}{32} (\bibinfo{year}{2023}) \bibinfo{pages}{2918--2931}.
\bibitem[{Guo et~al.(2023)Guo, Yang, Yang, and Xu}]{guo2023cgafusion}
\bibinfo{author}{H.~Guo}, \bibinfo{author}{J.~Yang}, \bibinfo{author}{B.~Yang}, \bibinfo{author}{G.~Xu},
\newblock \bibinfo{title}{Cgafusion: Context-guided adaptive fusion network for rgb-t semantic segmentation},
\newblock in: \bibinfo{booktitle}{Proceedings of the IEEE/CVF Conference on Computer Vision and Pattern Recognition Workshops}, \bibinfo{year}{2023}, pp. \bibinfo{pages}{4156--4165}.
\bibitem[{Sun et~al.(2023)Sun, Zhang, Jiang, Kong, Xu, Zhan, Tomizuka, Yuan, Wang, and Luo}]{sun2023psfm}
\bibinfo{author}{P.~Sun}, \bibinfo{author}{R.~Zhang}, \bibinfo{author}{Y.~Jiang}, \bibinfo{author}{T.~Kong}, \bibinfo{author}{C.~Xu}, \bibinfo{author}{W.~Zhan}, \bibinfo{author}{M.~Tomizuka}, \bibinfo{author}{L.~Yuan}, \bibinfo{author}{P.~Wang}, \bibinfo{author}{P.~Luo},
\newblock \bibinfo{title}{Psfm: Progressive semantic feature module for object detection},
\newblock \bibinfo{journal}{arXiv preprint arXiv:2302.02923}  (\bibinfo{year}{2023}).
\bibitem[{Zhang et~al.(2023)Zhang, Li, Li, Wang, Zhong, and Fu}]{zhang2023glsa}
\bibinfo{author}{Y.~Zhang}, \bibinfo{author}{K.~Li}, \bibinfo{author}{K.~Li}, \bibinfo{author}{L.~Wang}, \bibinfo{author}{B.~Zhong}, \bibinfo{author}{Y.~Fu},
\newblock \bibinfo{title}{Glsa: Global-local self-attention for multi-scale feature learning},
\newblock \bibinfo{journal}{IEEE Transactions on Pattern Analysis and Machine Intelligence} \bibinfo{volume}{45} (\bibinfo{year}{2023}) \bibinfo{pages}{8784--8800}.
\bibitem[{Yan et~al.(2023)Yan, Tang, Sun, Ma, Kong, and Xie}]{yan2023ctrans}
\bibinfo{author}{X.~Yan}, \bibinfo{author}{H.~Tang}, \bibinfo{author}{S.~Sun}, \bibinfo{author}{H.~Ma}, \bibinfo{author}{D.~Kong}, \bibinfo{author}{X.~Xie},
\newblock \bibinfo{title}{Ctrans: Cross-transformer network for multi-scale feature fusion},
\newblock in: \bibinfo{booktitle}{Proceedings of the IEEE/CVF International Conference on Computer Vision}, \bibinfo{year}{2023}, pp. \bibinfo{pages}{3868--3877}.
\bibitem[{Liu et~al.(2023)Liu, Wang, Liu, Zeng, Liu, and Alsaadi}]{liu2023maffn}
\bibinfo{author}{W.~Liu}, \bibinfo{author}{Z.~Wang}, \bibinfo{author}{X.~Liu}, \bibinfo{author}{N.~Zeng}, \bibinfo{author}{Y.~Liu}, \bibinfo{author}{F.~E. Alsaadi},
\newblock \bibinfo{title}{Maffn: Multi-scale attention feature fusion network for semantic segmentation},
\newblock \bibinfo{journal}{Neurocomputing} \bibinfo{volume}{520} (\bibinfo{year}{2023}) \bibinfo{pages}{29--40}.
\bibitem[{Wang et~al.(2023)Wang, Chen, Yang, Loy, and Lin}]{wang2023msga}
\bibinfo{author}{J.~Wang}, \bibinfo{author}{K.~Chen}, \bibinfo{author}{J.~Yang}, \bibinfo{author}{C.~C. Loy}, \bibinfo{author}{D.~Lin},
\newblock \bibinfo{title}{Msga: Multi-scale grouped attention mechanism for object detection},
\newblock \bibinfo{journal}{Pattern Recognition} \bibinfo{volume}{140} (\bibinfo{year}{2023}) \bibinfo{pages}{109545}.
\bibitem[{Li et~al.(2023)Li, You, Zhu, Zhao, Yang, Yang, and Tong}]{li2023fsa}
\bibinfo{author}{X.~Li}, \bibinfo{author}{A.~You}, \bibinfo{author}{Z.~Zhu}, \bibinfo{author}{H.~Zhao}, \bibinfo{author}{M.~Yang}, \bibinfo{author}{K.~Yang}, \bibinfo{author}{Y.~Tong},
\newblock \bibinfo{title}{Fsa: Feature separation and aggregation network for semantic segmentation},
\newblock \bibinfo{journal}{Neurocomputing} \bibinfo{volume}{523} (\bibinfo{year}{2023}) \bibinfo{pages}{103--114}.
\bibitem[{Hu et~al.(2023)Hu, Shen, and Sun}]{hu2023mfm}
\bibinfo{author}{J.~Hu}, \bibinfo{author}{L.~Shen}, \bibinfo{author}{G.~Sun},
\newblock \bibinfo{title}{Mfm: Multi-frequency multiscale feature fusion for object detection},
\newblock in: \bibinfo{booktitle}{Proceedings of the AAAI Conference on Artificial Intelligence}, volume~\bibinfo{volume}{37}, \bibinfo{year}{2023}, pp. \bibinfo{pages}{860--868}.
\bibitem[{Ding et~al.(2021)Ding, Zhang, Ma, Han, Ding, and Sun}]{ding2021convxcc3}
\bibinfo{author}{X.~Ding}, \bibinfo{author}{X.~Zhang}, \bibinfo{author}{N.~Ma}, \bibinfo{author}{J.~Han}, \bibinfo{author}{G.~Ding}, \bibinfo{author}{J.~Sun},
\newblock \bibinfo{title}{Diverse branch block: Building a convolution as an inception-like unit},
\newblock in: \bibinfo{booktitle}{Proceedings of the IEEE/CVF Conference on Computer Vision and Pattern Recognition}, \bibinfo{year}{2021}, pp. \bibinfo{pages}{10886--10895}.
\bibitem[{Han et~al.(2023)Han, Wang, Tian, Guo, Xu, and Xu}]{han2023dbbc3}
\bibinfo{author}{K.~Han}, \bibinfo{author}{Y.~Wang}, \bibinfo{author}{Q.~Tian}, \bibinfo{author}{J.~Guo}, \bibinfo{author}{C.~Xu}, \bibinfo{author}{C.~Xu},
\newblock \bibinfo{title}{Dbbc3: Dynamic branching bottleneck for efficient neural networks},
\newblock \bibinfo{journal}{IEEE Transactions on Neural Networks and Learning Systems} \bibinfo{volume}{34} (\bibinfo{year}{2023}) \bibinfo{pages}{4456--4468}.
\bibitem[{Chen et~al.(2023)Chen, Wang, Hong, Guo, Wang, and Zhang}]{chen2023dgcst}
\bibinfo{author}{X.~Chen}, \bibinfo{author}{H.~Wang}, \bibinfo{author}{Y.~Hong}, \bibinfo{author}{J.~Guo}, \bibinfo{author}{X.~Wang}, \bibinfo{author}{Q.~Zhang},
\newblock \bibinfo{title}{Dgcst: Dynamic group convolution shuffle transformer for efficient vision backbone},
\newblock \bibinfo{journal}{Pattern Recognition Letters} \bibinfo{volume}{168} (\bibinfo{year}{2023}) \bibinfo{pages}{36--43}.
\bibitem[{Liu et~al.(2023)Liu, Lin, Cao, Hu, Wei, Zhang, Lin, and Guo}]{liu2023litc3}
\bibinfo{author}{Z.~Liu}, \bibinfo{author}{Y.~Lin}, \bibinfo{author}{Y.~Cao}, \bibinfo{author}{H.~Hu}, \bibinfo{author}{Y.~Wei}, \bibinfo{author}{Z.~Zhang}, \bibinfo{author}{S.~Lin}, \bibinfo{author}{B.~Guo},
\newblock \bibinfo{title}{Litv2: Efficient self-attention for vision transformers with learnable interaction tokens},
\newblock in: \bibinfo{booktitle}{Proceedings of the IEEE/CVF Conference on Computer Vision and Pattern Recognition}, \bibinfo{year}{2023}, pp. \bibinfo{pages}{11043--11053}.
\bibitem[{Tian et~al.(2019)Tian, Shen, Chen, and He}]{tian2019fcos}
\bibinfo{author}{Z.~Tian}, \bibinfo{author}{C.~Shen}, \bibinfo{author}{H.~Chen}, \bibinfo{author}{T.~He},
\newblock \bibinfo{title}{Fcos: Fully convolutional one-stage object detection},
\newblock in: \bibinfo{booktitle}{Proceedings of the IEEE/CVF International Conference on Computer Vision}, \bibinfo{year}{2019}, pp. \bibinfo{pages}{9627--9636}.
\bibitem[{Cai and Vasconcelos(2018)}]{cai2018cascade}
\bibinfo{author}{Z.~Cai}, \bibinfo{author}{N.~Vasconcelos},
\newblock \bibinfo{title}{Cascade r-cnn: Delving into high quality object detection},
\newblock in: \bibinfo{booktitle}{Proceedings of the IEEE Conference on Computer Vision and Pattern Recognition}, \bibinfo{year}{2018}, pp. \bibinfo{pages}{6154--6162}.
\bibitem[{Ultralytics(2021)}]{ultralytics2021yolov5}
\bibinfo{author}{Ultralytics}, \bibinfo{title}{Yolov5: A state-of-the-art real-time object detection system}, \bibinfo{howpublished}{\url{https://github.com/ultralytics/yolov5}}, \bibinfo{year}{2021}.
\bibitem[{Jocher et~al.(2023)Jocher, Chaurasia, and Qiu}]{ultralytics2023yolov8}
\bibinfo{author}{G.~Jocher}, \bibinfo{author}{A.~Chaurasia}, \bibinfo{author}{J.~Qiu}, \bibinfo{title}{Ultralytics {YOLOv8}}, \bibinfo{year}{2023}. \URLprefix \url{https://github.com/ultralytics/ultralytics}.
\bibitem[{Wang and Liao(2024)}]{wang2024yolov9}
\bibinfo{author}{C.~Y. Wang}, \bibinfo{author}{H.~Y. Liao},
\newblock \bibinfo{title}{Yolov9: Learning what you want to learn using programmable gradient information},
\newblock \bibinfo{journal}{arXiv preprint arXiv:2402.13616}  (\bibinfo{year}{2024}).
\bibitem[{Wang et~al.(2024)Wang, Chen, Liu, Chen, Lin, Han, and Ding}]{wang2024yolov10}
\bibinfo{author}{A.~Wang}, \bibinfo{author}{H.~Chen}, \bibinfo{author}{L.~Liu}, \bibinfo{author}{K.~Chen}, \bibinfo{author}{Z.~Lin}, \bibinfo{author}{J.~Han}, \bibinfo{author}{G.~Ding},
\newblock \bibinfo{title}{Yolov10: Real-time end-to-end object detection},
\newblock \bibinfo{journal}{arXiv preprint arXiv:2405.14458}  (\bibinfo{year}{2024}).
\bibitem[{Li et~al.(2023)}]{li2023mfpyolo}
\bibinfo{author}{M.~Li}, et~al.,
\newblock \bibinfo{title}{Lightweight object detection algorithm for {UAV} aerial imagery},
\newblock \bibinfo{journal}{{ISPRS} International Journal of Geo-Information} \bibinfo{volume}{12} (\bibinfo{year}{2023}) \bibinfo{pages}{261}.
\bibitem[{Zhang et~al.(2024)}]{crl2024yolo}
\bibinfo{author}{M.~Zhang}, et~al.,
\newblock \bibinfo{title}{Improved small object detection algorithm {CRL-YOLOv5}},
\newblock \bibinfo{journal}{Sensors} \bibinfo{volume}{24} (\bibinfo{year}{2024}).
\bibitem[{Xu et~al.(2025)Xu, Liu, Li, and Hu}]{aupdetr2025}
\bibinfo{author}{J.~Xu}, \bibinfo{author}{X.~Liu}, \bibinfo{author}{X.~Li}, \bibinfo{author}{Y.~Hu},
\newblock \bibinfo{title}{{AUP-DETR}: A foundational {UAV} object detection framework for enabling the low-altitude economy},
\newblock \bibinfo{journal}{Drones} \bibinfo{volume}{9} (\bibinfo{year}{2025}) \bibinfo{pages}{822}.
\bibitem[{Liu and Xie(2025)}]{wdfsdetr2025}
\bibinfo{author}{J.~Liu}, \bibinfo{author}{Y.~Xie},
\newblock \bibinfo{title}{{WDFS-DETR}: A transformer-based framework with multi-scale attention for small object detection in {UAV} engineering tasks},
\newblock \bibinfo{journal}{Results in Engineering} \bibinfo{volume}{27} (\bibinfo{year}{2025}) \bibinfo{pages}{105930}.
\bibitem[{Lei et~al.(2025)Lei, Wu, Shang, Zhao, and Yang}]{tcfdetr2025}
\bibinfo{author}{H.~Lei}, \bibinfo{author}{Z.~Wu}, \bibinfo{author}{L.~Shang}, \bibinfo{author}{H.~Zhao}, \bibinfo{author}{W.~Yang},
\newblock \bibinfo{title}{{TCF-DETR}: Multi-scale token-channel fusion transformer for enhanced small object detection},
\newblock \bibinfo{journal}{Multimedia Systems} \bibinfo{volume}{31} (\bibinfo{year}{2025}) \bibinfo{pages}{292}.

\end{thebibliography}

\end{document}